\documentclass{article} %
\usepackage{iclr2027_conference,times}

\usepackage{amsmath,amsfonts,bm}

\def\eqref#1{equation~\ref{#1}}

\def\1{\bm{1}}

\DeclareMathAlphabet{\mathsfit}{\encodingdefault}{\sfdefault}{m}{sl}
\SetMathAlphabet{\mathsfit}{bold}{\encodingdefault}{\sfdefault}{bx}{n}

\usepackage{hyperref}
\hypersetup{pdftitle={HapBO: Harnessed α-π Bayesian Optimization via Expert Natural Language Priors},
            pdfauthor={Jean-S\'{e}bastien Giroux, Rose Guay Hottin, Marco Bonizzato}}
\usepackage{url}
\usepackage{xcolor}
\usepackage{fancyvrb} %
\usepackage{algorithm} %
\usepackage{booktabs}
\usepackage{algpseudocode}
\usepackage{graphicx}
\usepackage{amsmath}
\usepackage{amsthm}
\usepackage{amssymb}
\usepackage{float}
\usepackage{multirow}
\usepackage{enumitem}

\newcommand{\hapbo}{\textsc{HapBO}}
\newcommand{\apbo}{\textsc{$\alpha$-$\pi$BO}}
\newcommand{\ours}{\hapbo}

\newtheorem{theorem}{Theorem}
\newtheorem{proposition}{Proposition}
\newtheorem{lemma}{Lemma}
\newtheorem{definition}{Definition}
\newtheorem{remark}{Remark}
\newtheorem{assumption}{Assumption}
\newtheorem{corollary}{Corollary}

\title{\hapbo: Harnessed $\alpha$-$\pi$ Bayesian Optimization via Expert Natural Language Priors}

\author{%
  Jean-Sébastien Giroux$^{1,2}$, Rose Guay Hottin$^{1,2}$, Marco Bonizzato$^{1,2,3}$ \\[4pt]
  \begin{minipage}{1.15\textwidth}
  \raggedright\small
  $^1$Department of Electrical Engineering, Polytechnique Montréal, Montréal, Quebec, Canada.\\
  $^2$MILA -- Québec AI Institute, Montréal, Quebec, Canada.\\
  $^3$Department of Neurosciences, Centre Interdisciplinaire de Recherche sur le Cerveau et l'apprentissage (CIRCA) and Centre d'Innovation Biomédicale, Université de Montréal, Montréal, Quebec, Canada.
  \end{minipage}\\[4pt]
  Correspondence: \texttt{marco.bonizzato@polymtl.ca}
}

\iclrfinalcopy
\begin{document}

\maketitle

\begin{abstract}
Bayesian Optimization (BO) can be substantially more sample-efficient when informed by prior knowledge about the problem. In practice, however, such knowledge is often available only in natural language, e.g., expert intuitions or documentation, rather than as the explicit probability distribution landscape that prior-guided BO methods may require. Leveraging it effectively demands both an expressive mapping from language to priors and robustness to information that may be misleading. We propose \hapbo{} (Harnessed $\alpha$-$\pi$ Bayesian Optimization), which elicits beliefs once from a human expert, a large language model (LLM), or both, and deterministically compiles a structured specification into a reusable belief function ($\pi: \mathcal{X} \to[0,1]$), with a rich representation not restricted to a Gaussian family. Building on $\alpha$-$\pi$BO, \hapbo{} applies this belief function through a bounded positive multiplier on any nonnegative acquisition, preserving positive acquisition support and requiring no LLM calls during optimization. We prove that, for the Expected Improvement acquisition function (EI), \hapbo{} preserves the standard simple-regret rate up to a bounded factor, even under adversarial beliefs. We evaluate \hapbo{} on 40 HPOBench tasks, a synthetic suite in 6–15 dimensions and two real-world applications: cortical neurostimulation and constrained solar design. \hapbo{} achieves the lowest aggregate HPOBench regret and strong real-world task performance, with zero in-loop token cost. In real-world applications it improves the optimization outcome even at budgets of 12–20 queries, and at a small fraction of the wall-clock and token costs of other LLM-based methods.
\end{abstract}

\section{Introduction}
\label{sec:introduction}

Bayesian optimization (BO) is widely used for optimizing expensive black-box functions. An evaluation may for instance require patient time, hours of computation, or a physical experiment. A surrogate model, typically a Gaussian process (GP), approximates the objective, and an acquisition function selects the next query~\citep{MOCKUS1975428,jones1998efficient,frazier2018tutorialbayesianoptimization}. In most real-world scenarios, BO never truly starts the search from zero, since the practitioner usually has additional knowledge about the problem. For example, a machine learning engineer knows the plausible range of each hyperparameter of the network she is designing~\citep{eggensperger2022hpobenchcollectionreproduciblemultifidelity}, a neuroscientist knows which nervous system region controls a movement~\citep{bonizzato2023autonomous}, and a solar thermal engineer can have an estimation of which design parameters must be balanced against capital cost~\citep{andres-thio_solar_2024}. Used well, such knowledge buys the same result in fewer evaluations, and the gain is largest where evaluations are scarcest: at small budgets, where the surrogate alone (the internal model used by BO) is not mature yet~\citep{hvarfner2022piboaugmentingacquisitionfunctions, GUAYHOTTIN2025113039}. In real BO applications, such knowledge is most naturally expressed in language rather than as a probability distribution: an expert can describe where good configurations lie, and an LLM can produce a comparable description from a task specification alone. 

To take advantage of the prior belief, prior-guided BO techniques have been developed. These techniques allow an expert to inject a prior over the location of the optimum of the objective function. Since experts can make mistakes, these techniques may sometimes incorporate mechanisms that limit the damage when the prior is misleading~\citep{hvarfner2022piboaugmentingacquisitionfunctions, GUAYHOTTIN2025113039, 10.1007/978-3-030-86523-8_17, ICLR2024_b95f3f8b}. However, these techniques assume the prior is already available in numerical form, which practitioners are rarely equipped to provide. 

Here, our fundamental question is whether LLMs can bridge the gap between knowledge in its many native forms (e.g., a sentence by an expert, knowledge acquired during pretraining, or documents retrieved by web search) and a compiled belief function that is directly compatible with the landscape of GP models and BO acquisition functions. 

Large Language Models (LLMs) trained on internet-scale datasets have demonstrated broad knowledge and reasoning capabilities, so the BO community has tried to incorporate them within the BO loop. Current LLM-guided BO algorithms may use LLMs to 1) warm-start the search, 2) narrow the search space, or 3) replace core BO components (surrogate model or the acquisition function) during optimization~\citep{liu2024largelanguagemodelsenhance, ijcai2025p553, yuan2026unleashingllmsbayesianoptimization}. Our working hypothesis is that the GP-BO machinery is already well suited to provide high performance in most optimization tasks, from low to high dimension~\citep{hvarfner2024vanilla}. Moreover, in-loop LLM methods may incur substantial per-iteration cost, while lacking the calibrated uncertainty mechanisms developed for prior-guided BO techniques~\citep{kristiadi2024soberlookllmsmaterial, gupta2025llmsbayesianoptimizationscientific}. Methods that elicit the belief once avoid this per-iteration cost, but most use it in one of two limited ways: to warm-start the search, so that the belief shapes only the first few evaluations~\citep{liu2024largelanguagemodelsenhance}, or to prune the search space, so that any region it excludes can never be revisited~\citep{Lao_2024}. The first may discard the belief too early; the second may trust it too much. \cite{patel2026llmei} distill the LLMs knowledge once into a utility function, learned from its stated preferences, that guides BO toward promising regions. \ours{} shares this elicit-once design, but compiles the belief deterministically from a typed language and injects it through a bounded multiplier. 

\begin{figure}[h]
\centering
\includegraphics[width=0.95\linewidth]{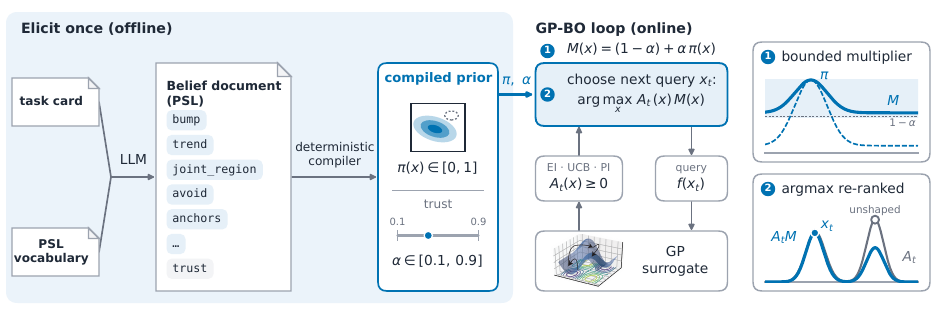}
  \caption{\textbf{\hapbo{} (blue components).}  Left to right: an LLM reads the task card, which may include information stated by a human expert, and the
  PSL vocabulary and produces a document in the typed Prior Specification Language
  (PSL); a compiler turns the document into a prior landscape
  $\pi$ and a trust level $\alpha$.  The prior is injected in the BO loop at the acquisition function level.  Right: shaping re-ranks candidates
  toward plausible regions but cannot zero a positive base acquisition.}
\label{fig:method}
\end{figure}

To overcome these gaps, we introduce \hapbo{} (Harnessed $\alpha$-$\pi$ Bayesian Optimization). \hapbo{} elicits beliefs once from a human expert, an LLM, or both, and deterministically compiles a structured specification into a  belief function ($\pi$: $\mathcal{X}$ $\to[0,1]$;  Fig.~\ref{fig:method}). Our contributions are as follows:
\begin{enumerate}[label=\textbf{(\roman*)}, leftmargin=*, itemsep=1pt, topsep=2pt, parsep=0pt]
    \item To turn natural-language knowledge into a BO-compatible landscape, we designed the Prior
  Specification Language (PSL, v1.0), a vocabulary built for versatility and completeness:
  its layered tokens express the structure experts may want to indicate (baseline trends, marginal
  effects, pairwise interactions), and anchors let it approximate any landscape to arbitrary precision 
  (Theorem~\ref{thm:psl-expressivity}). A PSL document compiles deterministically into a landscape, which is also
  compatible with traditional prior-injection methods (a prior on the GP mean or on the
  acquisition function).
    \item We choose to build upon $\alpha$-$\pi$BO, which allows \hapbo{} to apply this belief function through a bounded positive multiplier on any nonnegative acquisition, preserving positive acquisition support and requiring no LLM calls during optimization. We prove that, for the Expected Improvement acquisition function (EI), \hapbo{} preserves the standard simple-regret rate up to a bounded factor, even under adversarial beliefs.
    \item To evaluate \hapbo{} in practice, we ask whether \hapbo{} can improve sample efficiency of State-Of-The-Art (SOTA) methods: 1) one-shot LLM strategies that warm-start or constrain the search space, 2) methods that keep the LLM inside the BO loop, and 3) traditional prior-free search. We specifically ask how robust \hapbo{} is against misleading LLM-elicited beliefs, a potential roadblock that its component $\alpha$-$\pi$BO is designed to alleviate.
    \item We also present a series of ablations to interrogate the value of each component equipped to \hapbo{} by our chain of design choices. Our questions include whether results vary with the injection site, the belief representation, and the source of the belief.
\end{enumerate}

\section{Related Work}
\label{sec:related_works}

\textbf{Prior-guided BO.} A substantial corpus of work focuses on allowing practitioners to inject a prior over the location of the optimum. 
The resulting methods differ in how they limit the damage of a misleading belief, which they do in one of four ways.

First, by a schedule. $\pi$-BO weights the acquisition function by the prior landscape $\pi$, raised to an
exponent that decays with the number of iterations, regardless of the observed
data~\citep{hvarfner2022piboaugmentingacquisitionfunctions}. Second, by accumulating evidence. The prior's influence shrinks as observations contradict it, either
through the model, as in \textsc{BOPrO}, which combines the prior with a
data-driven model into a pseudo-posterior~\citep{10.1007/978-3-030-86523-8_17},
and \textsc{ColaBO}, which conditions the Gaussian process (GP) on the
belief~\citep{ICLR2024_b95f3f8b}, or through the sampling policy, as in
\textsc{PriorBand}, which adapts its reliance on prior-based samples in
multi-fidelity hyperparameter optimization (HPO)~\citep{NEURIPS2023_1704fe7a},
and \textsc{HypBO}, which distributes effort between expert-specified hypothesis
regions and the full space~\citep{ijcai2024p429}.
Third, by explicitly testing the belief. \textsc{DynaBO} combines decay with a  surrogate-based safeguard test that rejects flagged
priors~\citep{b1fd3044b2ec499387994421d5c6e938}, and \textsc{PSH} proves a no-harm
guarantee for prior-guided successive halving~\citep{hennig2026provablyreducedsamplecost}.
Fourth, by bounding the prior's influence structurally. \apbo{} (although descending from $\pi$-BO) does not insert a fading schedule, but rather places a fixed floor on the acquisition value, which results in quickly incorporating ``good'' priors, while drowning misleading ones, without a decay schedule and without deciding whether the belief is wrong~\citep{GUAYHOTTIN2025113039}. 
We build the \hapbo{} framework on the latter mechanism, but we modify the normalization of this construction to maintain a direct correspondence with our theoretical formulation.

A common challenge across all methods, model-based and acquisition-based alike, is that they require beliefs in a formal probabilistic form that the surrogate or the acquisition function can absorb. We here add the language interface that produces it.

\textbf{LLMs for BO.} Another line of work incorporates LLMs within the BO loop. These approaches vary mainly in the frequency with which the LLM is queried during optimization. Some approaches query the LLM before the search as a warm-starting strategy, either to select initial points to initialize the surrogate model~\citep{liu2024largelanguagemodelsenhance, ijcai2025p553}, or to produce a prior that persists throughout the search~\citep{Lao_2024}. Others query the LLM at every iteration of the optimization by using LLMs to replace core components of the BO loop such as the surrogate model or the acquisition function (AF), among others~\citep{liu2024largelanguagemodelsenhance, NEURIPS2025_c03a2610, chen2025informingacquisitionfunctionsfoundation,  yuan2026unleashingllmsbayesianoptimization, ICLR2025_a79237d6}, or query the LLM on demand to guide the AF in the selection of the next point to sample~\citep{ ijcai2025p553,  chen2026labollmacceleratedbayesianoptimization}. Our working hypothesis is that warm-starting initial points alone may not exploit the full capacity of BO priors. Also, available methods that produce priors that persist during the search may miss a key mechanism for exploiting it effectively, and in consequence can stop their action at producing the prior, and use it to constrain the search space rather than to weight it. Finally, using LLMs in the loop introduces additional challenges of uncalibrated uncertainty, volatility, which can cause having these methods consistently outperformed by classical surrogate-based methods. They also incur a query cost for knowledge that helps only in domain-specific settings~\citep{kristiadi2024soberlookllmsmaterial, gupta2025llmsbayesianoptimizationscientific, chang2025llinbotrustworthyllmintheloopbayesian, kobalczyk2026lilobayesianoptimizationnatural}. \hapbo{} therefore proposes to query the LLM before the search, leaving core BO components unchanged. Rather than warm-starting the surrogate model, or constraining the search, we give the acquisition function a prior to use through the entire search.

\textbf{Belief elicitation.} Outside of BO, \textsc{AutoElicit} generates Gaussian priors over the parameters of a linear model~\citep{capstick2025autoelicitusinglargelanguage}. Also, \textsc{LLM-Prior} converts rich, unstructured context, e.g. natural-language descriptions, data, or figures into analytically convenient probability distributions suitable for Bayesian inference ~\citep{huang2025llmpriorframeworkknowledgedrivenprior}. \hapbo~differs from these methods, since both stop at the prior, whereas ours is carried into the BO loop.

\section{Methodology}
\label{sec:methodology}

\subsection{Belief elicitation}
\label{subsec:belief_elicitation}

The first step of \hapbo{} is eliciting a belief in natural language. This stage requires a prompt to an LLM. The second step is to provide the vocabulary to the LLM to transcribe the elicited belief into a document that is then compiled deterministically. Two cards are used to drive this step: the task card provides the LLM's context, and the vocabulary card harnesses what the belief document can contain.

\textbf{Task card.} The task card gives the context needed by the LLM to write the belief document. At minimum, it shall contain three sections: the domain, which describes the system under study; the search space, which specifies each parameter; and a description of the optimization problem. The card can further contain human expert knowledge, pertinent data, and a scope of the search process (e.g., deep research, looking for literature sources, access to data). The design of the task card is open to the practitioner implementing \hapbo{} in their research, and benefits from prompt engineering skills developed with common LLM use.
In one experiment, we distinguish between cards written from three knowledge sources, whose variants differ only in what the card contains. \textit{LLM-only} cards ask the LLM to act as a domain expert and form the belief from its own knowledge; \textit{human-expert} cards carry substantial expert's belief, including available data from previous experiments; and \textit{mixed} cards carry the expert's belief written in their own words and ask the LLM to complement it with its own knowledge. 

\textbf{Vocabulary card.} The vocabulary card is fixed by \hapbo{} design, and defines the Prior Specification Language (PSL, v1.0, see Appendix~\ref{app:psl}), a closed grammar for writing beliefs, and restricts the LLM to translation: the LLM writes the belief in PSL and never computes a prior. PSL expresses per-dimension shapes, pairwise interactions, penalized regions, and anchor points. The resulting belief is joint over the search space and not restricted to a Gaussian family.

\textbf{Belief document.} Both cards are sent together in a single LLM call, which returns a belief document containing the typed search space, the list of beliefs, and a qualitative trust block. Each shaping belief carries a weight (its relative importance) and a provenance tag (its source), both used by the compiler (see Section~\ref{subsec:det_compiler}). An example of an elicited document is given in Appendix~\ref{app:psl}.

\subsection{Deterministic compilation}
\label{subsec:det_compiler}

The compiler maps the belief document to the two quantities used by \apbo{}: a prior landscape $\pi:\mathcal{X}\to[0,1]$ and a trust level $\alpha \in [0.1, 0.9]$ (where a higher value means more trust in the prior). The compiler first validates the belief document against the PSL vocabulary, rejecting any unknown PSL token, field, or out-of-range value. It then translates each token into a predefined formula (see Table~\ref{tab:psl} and Algorithm~\ref{alg:compile}, Appendix~\ref{app:psl}), and computes $\alpha$ from the trust block and the provenance tags by a fixed rule (Eq.~\ref{eq:alpha}). It involves no LLM call and no randomness, so the same belief document always yields the same prior. 

\textbf{Building the prior $\pi$.} Each PSL token has a predefined closed-form shape on the search space, encoded to the unit cube so that all dimensions share the same scale (Table~\ref{tab:psl}, Fig.~\ref{fig:psl_components}). The compiler proceeds to build the prior by sequentially adding the weighted shapes to a baseline, subtracting penalties, clamping the sum at zero, multiplying it by the preference gates, and rescaling it to $[0,1]$. The anchors are then used to set exact specified values at the anchor points. Eqs.~(\ref{eq:score})--(\ref{eq:anchors}) and Algorithm~\ref{alg:compile} in Appendix~\ref{app:psl} give the full procedure; Fig.~\ref{fig:compile-trace} visualizes the compiler algorithm.

\textbf{Determining the trust level $\alpha$.} We opted for an automatic simplistic rule, that allows us to illustrate two possible practitioner choices at once. We combine the document's LLM-stated confidence $c_{\text{overall}}$ with the provenance of each belief,
\begin{equation}
\alpha=\operatorname{clip}\!\left(c_{\text{overall}}\cdot\langle c_{\text{provenance}}\rangle_w,\;0.1,\;0.9\right),
\label{eq:alpha}
\end{equation}
where $\langle\cdot\rangle_w$ is the average weighted by the belief weights. A higher $\alpha$ means more trust in the prior.  Capping $\alpha$ at $0.9$ bounds the worst-case constant of our guarantee at $10$ (Proposition~\ref{prop:shaping} in App.~\ref{app:theory-shaping}). 
In our implementation, we set the belief stated by the expert to a higher trust level than one supplied by the LLM. This idea is set to showcase at once the combination of an LLM-stated confidence, with a more `human' fixed trust rule. Nevertheless, a practitioner using \hapbo{} could effectively substitute our simplified $\alpha$-determination module with an LLM decision-maker, or even with a human-tuned cursor.

\subsection{Shaping the acquisition function}
\label{subsec:af_shaping}

\hapbo{} leaves the BO loop (presented in Appendix~\ref{app:gpbo}) unchanged except for the final ranking. At each query it selects
\begin{equation}
x_t\in\arg\max_{x\in\mathcal X}A_t(x)\,M(x),\qquad M(x)=(1-\alpha)+\alpha\,\pi(x),
\label{eq:hapbo}
\end{equation}
where $A_t\geq0$ is the raw base acquisition, e.g., EI: 
\begin{equation}
 A_t(x)=\bigl(\mu(x)-\hat y\bigr)\,\Phi\bigl(z(x)\bigr)+\sigma(x)\,\phi\bigl(z(x)\bigr)
\end{equation}
with
  $z(x)=(\mu(x)-\hat y)/\sigma(x)$; (App.~\ref{app:gpbo}). 

  The prior landscape $\pi$ and the trust $\alpha$ are compiled once before the run (Section~\ref{subsec:det_compiler}).
  
 Following \citet{GUAYHOTTIN2025113039}, the belief acts as a bounded multiplier (which their experiments found to be more robust to misleading priors than $\pi$-BO); unlike their formulation, which min--max normalizes the acquisition before multiplying, we apply $M$ to the raw acquisition, maintaining a direct correspondence with our theoretical formulation (Definition~\ref{def:shaping} in App.~\ref{app:theory}).
When comparing with Vanilla BO, the surrogate, the base acquisition, and the candidate set are kept identical, and both start from the same random point. Because the belief enters only through $M$, the rule applies to any nonnegative acquisition (Appendix~\ref{app:variants}).

\subsection{Theory guarantees}
\label{subsec:theo_guarantees}

\textbf{The PSL harness} itself is free of expressive bottlenecks, because 1) anchors-only documents can reproduce any
belief table on a finite candidate set exactly, and 2) approximate any continuous
belief uniformly with error $O(K^{-1/d_c})$ in the number of anchors $K$ over $d_c$
continuous dimensions (Theorem~\ref{thm:psl-expressivity}).

 \textbf{Regarding the $\alpha$-$\pi$ harness}, since $\alpha\leq0.9$, the multiplier satisfies $0.1\leq M(x)\leq1$, so no candidate with positive acquisition is ever ruled out, and the selected point is always within a factor $10$ of the raw acquisition's maximum (Proposition~\ref{prop:shaping}). For noiseless Expected Improvement with a fixed kernel, shaping preserves the classical simple-regret rate of \citet{JMLR:v12:bull11a} up to a constant $1/(1-\alpha\beta_\star)\leq10$, where $\beta_\star=1-\pi(x^\star)$ is the prior's miss at the optimum (Theorem~\ref{thm:ei}, Corollary~\ref{cor:ei-scalar}); a prior that is correct at the optimum costs nothing, however wrong it is elsewhere. For noisy GP-UCB, the regret bound gains an additive term that depends on the belief only through its value at the optimum (Theorem~\ref{thm:ucb}).

Full statements, assumptions, and proofs are given in Appendix~\ref{app:theory}.

\section{Experimental setup}
\label{sec:experiments}

\textbf{Benchmarks.} HPOBench contributes 40 tabular hyperparameter-optimization tasks spanning five learner families~\citep{eggensperger2022hpobenchcollectionreproduciblemultifidelity}.   The  NeuroMOSAICS Cortical panel optimizes evoked muscle responses across 40 rat-M1 electrode tasks (eight hindlimb muscles recorded in six rats) \citep{kardassevitch2025neuromosaics}. SOLAR MAXEFF\_RE is a constrained solar-design black box \citep{andres-thio_solar_2024}.
We further designed an analytic panel, containing high-dimensional synthetic functions: Hartmann-6, Styblinski--Tang-7, Levy-10, and Ackley-15~\citep{https://doi.org/10.1002/nav.3800200316, styblinski1990experiments, levy1985tunneling, ackley2012connectionist}. For this panel, we generate three controlled prior fidelities per experiment, by manipulating the synthetic functions to create alternate functions within the same family but having a diverse landscape (correlation $\rho = 0.1$  to $0.9$, from poor to accurate). We then feed the manipulated function to the elicitor, informing the LLM that this is our expert belief for the desired landscape. The elicited prior is tested against the true synthetic function, with shifted optima to exclude possible memorization effects (e.g. Ackley optimum is no longer at the origin). We further discuss LLM memorization leak control in Appendix~\ref{app:leak-probe}.  

\textbf{Baselines.} We compare \hapbo{} against an array of  methods. Some are \emph{prior-free}: vanilla BO, random search and \textsc{TuRBO}~\citep{NEURIPS2019_6c990b7a}. Other methods are \emph{prior-guided}, like  \textsc{ColaBO}~\citep{ICLR2024_b95f3f8b}, which conditions the GP posterior on a user belief, which we provide via an LLM query following \textsc{BORA}'s LLM-prior recipe~\citep[supplement C.2]{ijcai2025p553}, and we particularly focus on other methods incorporating a \emph{LLM-in-the-loop} approach: \textsc{LLAMBO}~\citep{liu2024largelanguagemodelsenhance}, which uses the LLM as surrogate and candidate sampler; \textsc{BORA}~\citep{ijcai2025p553}, which lets the LLM propose regions during the search; and \textsc{LGBO}~\citep{yuan2026unleashingllmsbayesianoptimization}, which shifts the GP mean by an LLM preference at each query. The in-loop methods receive the same task card as \hapbo{}, through their own interface, and query the same locally served LLM. Where a baseline admits more than one faithful implementation, we report the strongest variant we could construct (Appendix~\ref{app:baseline-impl}). Further methods, \textsc{CAKE}~\citep{NEURIPS2025_c03a2610}, \textsc{CALIPER}~\citep{chen2026evidencegatedllmpriorsmultiobjective}, \textsc{PriorBand}~\citep{NEURIPS2023_1704fe7a}, \textsc{PriMO}~\citep{basu2025primo}, \textsc{HypBO} (fed LLM hypotheses)~\citep{ijcai2024p429} , and \textsc{LLM-EI}~\citep{patel2026llmei}, are reported in Appendix~\ref{app:cake}.

\textbf{Protocol.} Every baseline runs with the same
locally served LLM (Qwen3.5-122B; \citealp{qwen3.5}) and the same task descriptions, served through its native interface. \hapbo{} and Vanilla BO share the surrogate, acquisition, candidate set, and initial point (same as random search), and differ only in the presence or absence of the prior. 

For \hapbo{}, priors are elicited once per task at temperature 0 for HPOBench, cortical and SOLAR, and as three independent single-call draws per card for the synthetic panel (default temperature 0.6): the elicitor receives a de-identified task card and must answer in the closed PSL vocabulary
(Appendix~\ref{app:psl}), with validator errors fed back for up to four
corrective turns (Appendix~\ref{app:cards}). Per-benchmark
normalizations, SOLAR's constraint handling, and per-query wall-clock and token costs are detailed in Appendix~\ref{app:details}. 

\textbf{Metrics.} We evaluate query-wise best-so-far performance. Simple regret is used when a reference optimum is known; SOLAR reports best feasible efficiency because its optimum is unknown. Means and uncertainty bands are computed over tasks or subjects after averaging 10 paired seeds.

\section{Results}
\label{sec:analysis}

\subsection{HPOBench and real-world tasks}

\begin{figure}[t]
\centering
\includegraphics[width=0.9\linewidth]{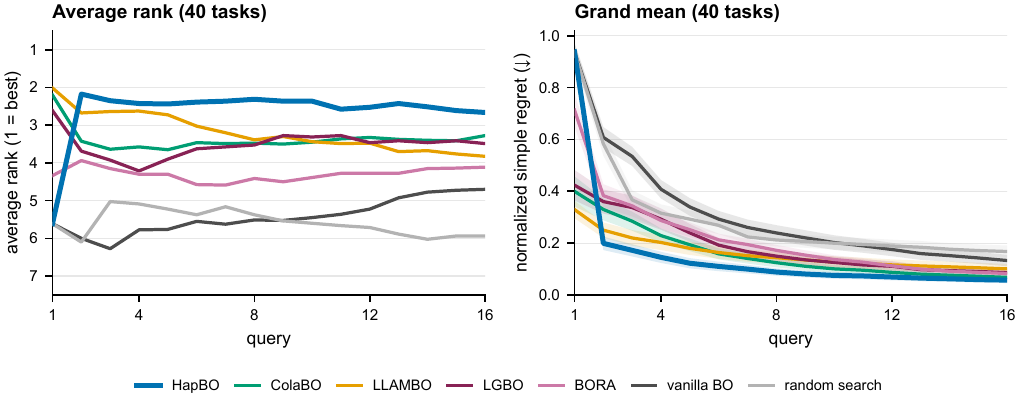}
\caption{\textbf{HPOBench.} Average rank and normalized simple regret as a grand mean over 40 tasks for five model families. 
Per-family panels in App.~\ref{app:hpo-family}; bands are $\pm$SEM across tasks after averaging seeds.}
\label{fig:hpo}
\end{figure}

We first compared \hapbo{} to the SOTA and traditional baselines on HPOBench. \hapbo{} has the lowest grand-average regret curve from the second query on 
(Fig.~\ref{fig:hpo}).  Its mean raw endpoint regret, averaged within and then across the five
model families, is $0.0026$, compared with $0.0039$ for ColaBO, $0.0145$ for its matched
random-search control, $0.0074$ for LLAMBO, $0.0092$ for its matched vanilla-BO control, $0.0053$
for LGBO, and $0.0048$ for BORA.  The rank
panel in Fig.~\ref{fig:hpo} positions \hapbo{} ahead of every baseline in the performance ranking across HPOBench problem families. 
Two further SOTA methods, \textsc{CAKE} and \textsc{CALIPER}, are competitive
under a faithful prior but do not overtake \hapbo{} (Appendix~\ref{app:cake},
Fig.~\ref{fig:cake-caliper}). \textsc{PriorBand} and \textsc{PriMO}
are less performant under the same good LLM-elicited prior and substantially on par under an adversarial prior (Fig.~\ref{fig:priorband-primo}). \textsc{HypBO} with LLM hypotheses and
\textsc{LLM-EI} also trail \hapbo{} under both faithful and adversarial priors
(Fig.~\ref{fig:hypbo-llmei}).

In two real-world experiments, pertaining to cortical neurostimulation and solar design, \hapbo{} is the most performant method (second best to LLAMBO on cortical). However, unlike LLAMBO, it never fails completion. Table~\ref{tab:endpoints} collects the endpoints.

\textbf{Cost.}  The efficiency gains are bought with modest compute overhead.  \hapbo{} pays the LLM once, at
setup (${\sim}15{,}000$ tokens to elicit and compile one HPO prior), then runs as a pure GP loop at
$0.20$\,s and zero tokens per query, the same as vanilla BO. Conversely, SOTA baselines have larger expenditures, with  LLAMBO spending
${\sim}79{,}700$ tokens and $28.6$\,s per query, ColaBO $104$\,s per query for its
prior-conditioned sampling. BORA and LGBO, which engage in more compute time per query, manage to mitigate the tokens/query figure, sitting at few hundreds of tokens per query on average
(App. Table~\ref{tab:cost}).

\begin{figure}[t]
\centering
\includegraphics[width=0.75\linewidth]{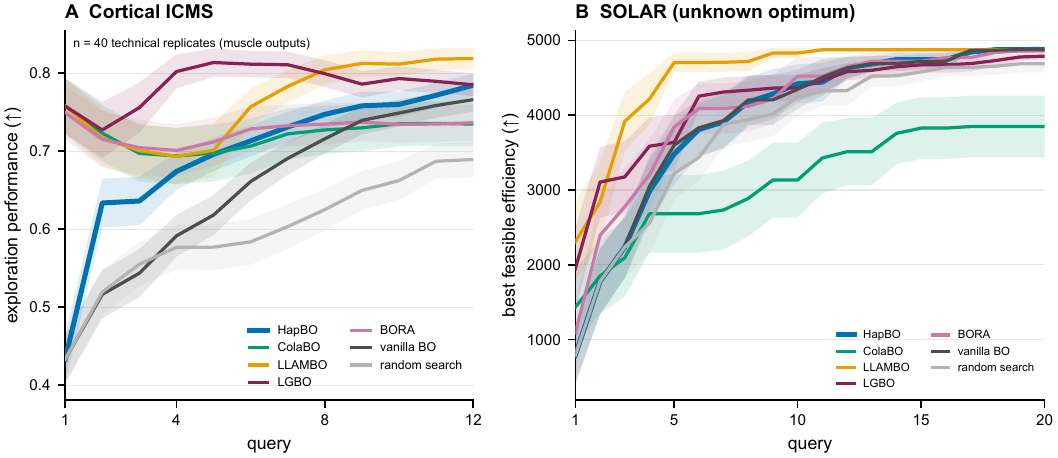}
\caption{\textbf{Real-world tasks.} \textbf{A:} NeuroMOSAICS cortical exploration performance (higher is better); n=40 technical replicates (muscle outputs), seeds averaged within each map.  \textbf{B:} SOLAR\_MAXEFF best
feasible efficiency; zero denotes that no feasible point has yet been found.}
\label{fig:real}
\end{figure}

\begin{table*}[t]
\centering
\caption{\textbf{Endpoint performance per task family.}  HPOBench simple regret for each of the five
model families (lr, rf, xgb, nn, svm; lower is better), cortical exploration performance
(higher is better --- the same unit as
Fig.~\ref{fig:real}A), and SOLAR best
feasible efficiency (higher); regrets are comparable across HPO families because all are validation
error.  \emph{fail\%} is the pooled fraction of task--seed runs that did not complete across every family
(HPO, cortical, and SOLAR).  \hapbo{} uses matched random initialization on HPO, cortical, and SOLAR
alike.  Bold marks the best value in each performance
column.}
\resizebox{\textwidth}{!}{\begin{tabular}{lcccccccc}
\toprule
Method & lr & rf & xgb & nn & svm & cortical & SOLAR & fail\%\\
\midrule
HapBO & 0.0022 & \textbf{0.0015} & \textbf{0.0020} & \textbf{0.0052} & \textbf{0.0021} & 0.785 & \textbf{4881.2} & 0.0\\
ColaBO & \textbf{0.0014} & 0.0031 & 0.0022 & 0.0102 & 0.0028 & 0.735 & 3848.9 & 0.0\\
LLAMBO & 0.0017 & 0.0201 & 0.0021 & 0.0090 & 0.0041 & \textbf{0.819}$^\dagger$ & 4874.9$^\dagger$ & 1.9\\
LGBO & 0.0023 & 0.0088 & \textbf{0.0020} & 0.0075 & 0.0060 & 0.785 & 4789.6 & 0.0\\
BORA & 0.0022 & 0.0074 & \textbf{0.0020} & 0.0081 & 0.0042 & 0.737 & 4858.9 & 0.0\\
vanilla BO & 0.0020 & 0.0243 & \textbf{0.0020} & 0.0119 & 0.0059 & 0.766 & 4877.9 & 0.0\\
random & 0.0029 & 0.0387 & 0.0023 & 0.0094 & 0.0191 & 0.690 & 4687.3 & 0.0\\
\bottomrule
\end{tabular}
}

{\footnotesize $^\dagger$~performance limited to successful runs, presented undiscounted for failures.} 
\label{tab:endpoints}

\end{table*}

\subsection{Robustness in the accurate-to-poor prior spectrum}

We next ask what price an adversarial misbelief would carry within the optimization process. An enticing feature of prior-based BO methods is the capacity to disregard prior information that is proven non-performant. Eventually, the ideal method would be expected to close the gap, query after query, with prior-free variants under these conditions. 

\begin{figure}[t]
\centering
\includegraphics[width=0.9\linewidth]{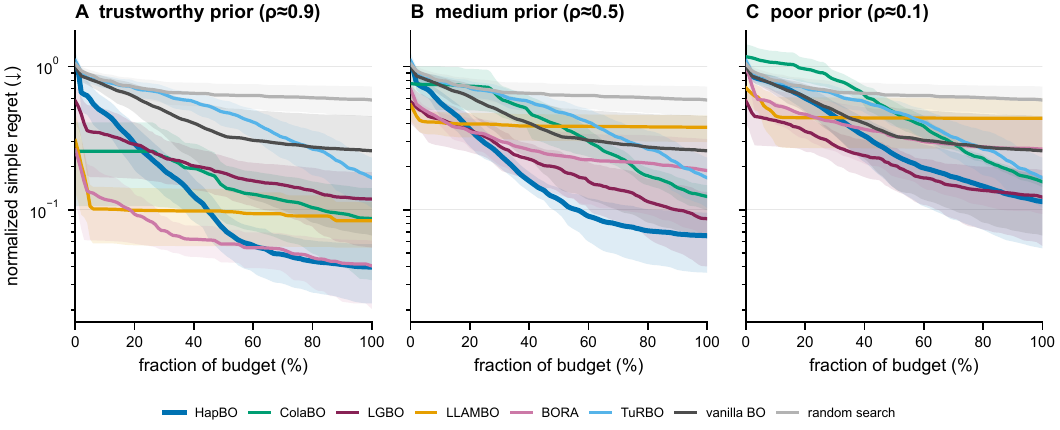}
\caption{\textbf{Synthetic prior-fidelity sweep.} Grand-average simple regret under trustworthy
($\rho\approx0.9$), medium ($\rho\approx0.5$), and poor ($\rho\approx0.1$) priors on 4 synthetic functions.  Each
function is evaluated at its own budget and aligned only by fraction of budget for this summary.}
\label{fig:synthetic}
\end{figure}

We also propose a synthetic function experiment, where the prior fidelity (as a modified function expressed in natural language with an equation) is controlled for exact correlation levels with respect to the true target function to be optimized. Our experiment makes the cost of prior
quality visible for every method on one display (Fig.~\ref{fig:synthetic}).  At all $\rho$
values, prior-using methods have an advantage with respect to prior-free controls (except LLAMBO, which seems to only offer acceptable performance at $\rho=0.9$). This advantage modulates with the quality of information received: it reduces as $\rho$ falls to $0.5$ and
$0.1$. Fig.~\ref{fig:synthetic} shows that  \hapbo{}'s performance is always best across all conditions. 

\begin{figure}[t]
\centering
\includegraphics[width=0.8\linewidth]{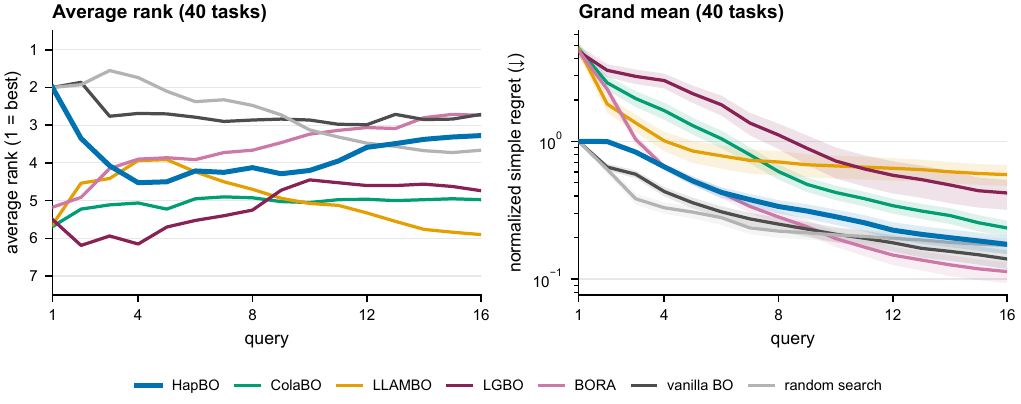}
\caption{\textbf{Adversarial HPOBench.}  Average task-wise rank (left; over the task
grid complete for every method) and grand-mean normalized simple regret over 40 tasks (right) when
every method receives the same hostile task description through its native interface.
Per-family panels in App.~\ref{app:adv-family}; bands are $\pm$SEM across tasks after averaging
ten seeds.}
\label{fig:robust}
\end{figure}

We then tested \hapbo{} and SOTA methods performance under matched adversarial priors for HPO (prior placed at each
task's empirical anti-optimum), with the same hostile description passed to every baseline.
\hapbo{} initially pays for the misleading belief but recovers better than any SOTA alternative (except BORA) toward the strongest
prior-free curves of random search (Fig.~\ref{fig:robust}).

\subsection{Elicitation modalities}

\textbf{Knowledge sources.} The same compiled interface accepts LLM-only, human-only, and combined beliefs. We show an example of this three-way usage of the LLM elicitor interface on the NeuroMOSAICS Cortical panel (Appendix~\ref{app:sources-full}).
We elicit the prior with 1) no hint (LLM only), 2) a modest injection of human-level
expertise in the form of a single informative sentence (LLM+human), and 3) substantial prior information in the form of previous
experimental data not included in the benchmark dataset (human only). The LLM+human and human-only priors take less than half the total number of queries to hit performance levels aligned with the final one of vanilla BO; the LLM-only prior does not improve on vanilla BO because of lacking domain knowledge (Fig.~\ref{fig:sources-full}G--I). Just a single line of human expert information thus improves \hapbo{} performance in this niche application.

\textbf{Elicitor LLM.} We further test different elicitor LLMs, including commercial frontier models
(App.~\ref{app:ladder}). Leveraging more potent models can result in a more accurate modeling of the target function by PSL, albeit with modest gains in regret in the BO phase.

 \subsection{Ablations}
 \label{sec:ablations}

\textbf{Injection site.}  In this ablation we ask whether changing the injection site for a prior leads to a difference in performance, as expected from literature~\citep{hvarfner2022piboaugmentingacquisitionfunctions, GUAYHOTTIN2025113039}. We fixed the representation and moved the site: 1) the bounded
acquisition multiplier of \ours{}, 2) the no-floor $\pi$-BO, 3) a prior to the mean of the surrogate
(prior-to-mean), or 4) the initial queries (prior-warm-start), with a prior-only (non-BO) greedy arm and
no-prior controls. Across HPOBench and the synthetic functions, \hapbo{} is the strongest landscaped method (vs prior-to-mean), with equal final performance to $\pi$-BO on good priors but more robust than $\pi$-BO on misleading or intermediate priors. $\pi$-BO is known to be less robust to low-fidelity priors across benchmarks~\citep{GUAYHOTTIN2025113039}. As expected, all landscaped algorithms avoid dramatic collapse of performance when the prior is misleading (Fig.~\ref{fig:ablations}). There, \hapbo{}'s loss is modulated by its assigned trust weight $\alpha$ (Fig.~\ref{fig:alpha-sweep}).

\begin{figure}[t]
\centering
\includegraphics[width=\linewidth]{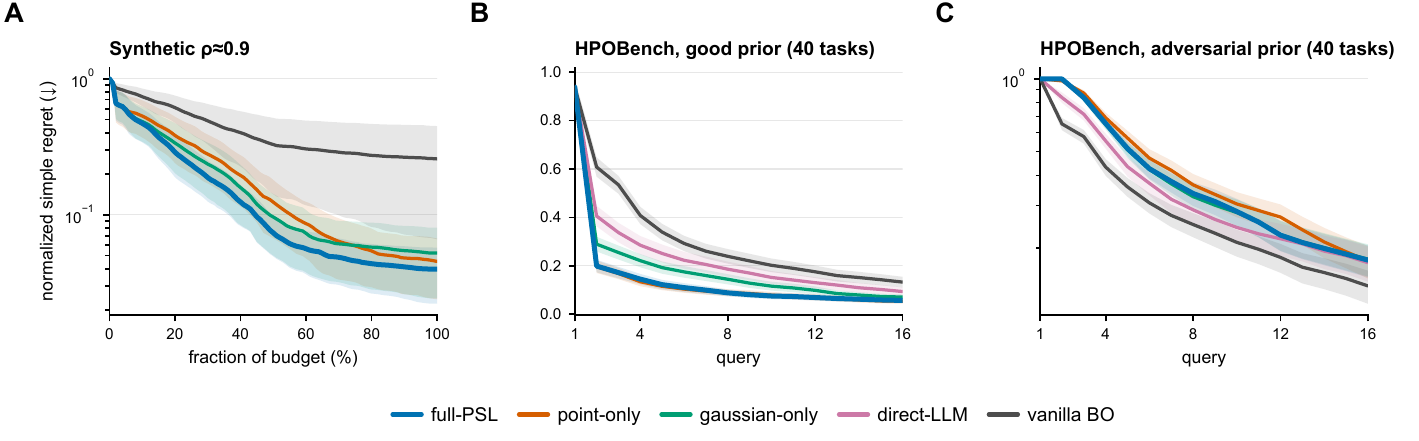}
\caption{\textbf{Prior representation ablation.}  The injection site is fixed at the bounded $\alpha-\pi$BO
multiplier; only how the belief becomes $\pi$ varies: full PSL (ours), point-only, Gaussian-only, direct-LLM elicitation,
and vanilla BO (no prior).  Grand-mean normalized simple regret on \textbf{A:} the synthetic suite at $\rho\approx0.9$
(four analytic functions, aligned by fraction of budget), \textbf{B:} HPOBench with the good prior, and \textbf{C:} HPOBench with the adversarial prior.
Direct-LLM is absent from the synthetic suite, because its fixed anchor set degenerates in 7--15D.  Ten seeds per task (synthetic prior arms: three documents $\times$ ten seeds per function); mean $\pm$ SEM across tasks; lower is better.  Average ranks and per-function and per-family panels in App. Fig.~\ref{fig:representation}.}
\label{fig:representation-main}
\end{figure}

\textbf{Representation.}  In this ablation we ask whether PSL holds a substantial advantage with respect to three more naive prior-building languages (point-only, direct LLM elicitation and gaussian-only). 
We show that all structured representations outperform direct LLM scoring when a good prior is available. Full PSL is best on the synthetic suite at $\rho=0.9$, and always aligned with the other best prior-based arms on the good prior (point-only) and adversarial prior (gaussian-only and direct-LLM).
Of note, direct-LLM elicitation in high-dimension PSL is not possible in cases such as our synthetic suite, since the fixed anchor set degenerates in 7-15D  (Fig.~\ref{fig:representation-main}; ranks and per-function and per-family panels in App. Fig.~\ref{fig:representation}).

\textbf{BO variants.} In this ablation we ask whether the bounded multiplier depends
on the BO machinery around it, since it acts only on the acquisition ranking. We
repeat the matched prior-on/prior-off contrast under EI, PI, and UCB, and under
sequential versus batch selection (Kriging Believer~\citep{ginsbourger2010kriging}, $q=3$), and three GP kernels. Across all three
acquisitions and kernels, shaping reduces final normalized HPOBench regret by $2.5$--$12\times$,
with the largest cortical gains at low budget, and a boost of early performance is present under
batch selection (Appendix~\ref{app:variants}, Fig.~\ref{fig:variants}).

\subsection{Knowledge leak control}

The objective of this paper is to determine, when LLM knowledge about a given problem exists,
whether this can be integrated, by proper harnessing, in BO. However, there is a type of existing
knowledge that is less desirable when testing any LLM-prior method on public benchmarks, which
is contamination: a strong prior might reflect recall of the final benchmark answer rather than trans-
ferable expertise. We address this with two complementary controls: a direct probe of what the
elicitor natively knows, and an additional objective whose optimum cannot have been memorized
(Appendix~\ref{app:leak}). 

First, we show that no
elicitor recalls an optimum on any real task: the blind guess recovers
$48$--$68\%$ of the head-room on HPOBench and falls below the map average on the
cortical grid, while naming a synthetic function outright recovers its optimum
exactly. Second, we build a bespoke CNN benchmark whose response surface is thus completely unseen to the model. There, \hapbo{} again reaches the lowest simple regret
(Fig.~\ref{fig:leak-control}).

\section{Conclusion and future work}
\label{sec:conclusion}
We presented \hapbo, a framework that takes natural-language beliefs, from a human expert, an LLM, or both, and converts them into a structured specification, which is then deterministically compiled into a reusable belief function with a rich representation palette. This belief function can be used by any BO method admitting a landscaped prior; we instantiate it with \apbo, which is robust to erroneous beliefs. Since the belief function is compiled once before optimization, the resulting pipeline requires no LLM calls during optimization, and rather admits a traditional (shaped) BO loop. Results on 40 hyperparameter optimization problems, a synthetic suite in 6--15 dimensions, and two real-world applications show that \hapbo~outperforms methods that keep the LLM in the loop under small budgets, in terms of performance, optimization time, and token cost; our experiments also show that \hapbo{} is very robust to misleading or poor priors, since such priors tend to only waste a small budget, and no region of the search space is ruled out by the prior-injection technique.  

Several limitations remain. First, the quality of the compiled belief depends on how information is presented to the LLM when converting natural language into a specification; more broadly, each component of the pipeline is modular and further research could improve some of our design choices. Elicitation design in particular is a promising direction for future work. In this regard, PSL was built for versatility and completeness: its layered vocabulary covers order-0, marginal and pairwise structure, and anchors make it dense in the space of landscapes (Theorem \ref{thm:psl-expressivity}). Future work could optimize PSL itself, for instance by refining its token set, the compiled shapes, or the guidance given to the elicitor, to further improve the toolbox available for elicitation.  Second, \hapbo~relies on one-shot elicitation: the belief function is compiled once before optimization and cannot be updated with what the search reveals. A natural extension is a hybrid approach in which the LLM occasionally intervenes during optimization to refine the belief. Finally, a misleading prior can still degrade performance relative to vanilla BO under small budgets; further limiting this negative effect is an additional  available direction for future work. 

Overall, we here advocate for a black-box optimization approach that uses the LLM offline to construct an auditable belief, then let an ordinary BO loop decide how much the data flow should trust or overturn it.

\subsection*{AI use statement}
\label{sec:ai_use_statement}

In this work, the authors used generative AI tools to formulate mathematical claims, provide critical ingredients for proving mathematical claims, assist in the writing of proofs, implement methods, and assist with translation.
The authors did not use generative AI tools to help develop theoretical models or conceptual frameworks, propose or refine hypotheses, design or provide feedback on research methodology or experiments, or interpret results. Supporting qualitative and thematic data analysis is not applicable to this work.
Additionally, the authors used generative AI tools to generate synthetic datasets (limited to the synthetic function experiment and the bespoke CNN experiment in appendix), reformat datasets (limited to SOLAR, allowing a tabular format for testing on an offline batch run), create or modify scientific figures or images, create or edit software code, draft parts of the paper, and edit the paper to improve readability. In each case, the authors verified that the scientific figures were accurate and free of errors, two authors verified and tested the code generated by AI coding agents, and the authors revised the drafted text. The authors take responsibility for the final content of this work, including any text, claims, or artifacts produced with the aid of generative AI.

\subsubsection*{Ethics statement}
The cortical experiments use a public rodent neurostimulation dataset
\citep{kardassevitch2025neuromosaics}; no animal experiments were conducted for this work.

\subsection*{Reproducibility statement}
\label{sec:reproducibility}

In order to make \hapbo{} results reproducible, we provide the source code for this project at \url{https://github.com/mbonizzato/HapBO}. There, readers have access to documentation with instructions on how to run each of the experiments presented in the main text and the appendix. We also provide readers with the pipeline that allows them to generate their own priors by modifying the appropriate \texttt{prompt\_card.md}. The data processing used for each dataset and instructions on how to download the datasets can be found in \texttt{benchmark/README.md}. For the theoretical results, we also provide explanations of any assumptions and a complete proof of the claims in the appendix.

\subsubsection*{Acknowledgments}
\label{sec:acknowledgments}
We thank the Digital Research Alliance of Canada (Calcul Québec) for providing computational resources for several arms of this study. This research was supported by the New Frontiers in Research Fund -- Transformation project RE-MOVE (NFRFT-2024-00209), the Natural Sciences and Engineering Research Council of Canada (NSERC; RGPIN-2023-04370), the equipment used was procured through a grant from Canada Foundation for Innovation.

\bibliography{iclr2027_conference}
\bibliographystyle{iclr2027_conference}

\appendix

\section{Gaussian-process Bayesian optimization}
\label{app:gpbo}

BO seeks to find $x^\star\in\arg\max_{x\in\mathcal X}f(x)$ for an expensive black-box function $f$ in as few
evaluations as possible \citep{MOCKUS1975428,jones1998efficient,frazier2018tutorialbayesianoptimization}.
Starting from initial observations $D_n=\{(x_i,y_i)\}_{i=1}^{n}$ with a possibly noisy $y_i=f(x_i)+\epsilon_i$, each
iteration 1) fits a probabilistic surrogate to $D_n$, 2) maximizes an acquisition function that scores
candidates by the surrogate's mean and uncertainty, 3) evaluates the maximizer, and 4) augments $D_n$. The incumbent $\hat y$ is the best response observed so far.  The surrogate throughout this paper
is a Gaussian process, $f\sim\mathcal{GP}(0,k)$: conditioned on $D_n$ under a Gaussian likelihood
with noise variance $\sigma^2_{\mathrm{lik}}$, the posterior at a test point $x$ is Gaussian with
\begin{equation*}
\mu(x)=\mathbf k_x^{\top}(K+\sigma^2_{\mathrm{lik}}I)^{-1}\mathbf y,\qquad
\sigma^2(x)=k(x,x)-\mathbf k_x^{\top}(K+\sigma^2_{\mathrm{lik}}I)^{-1}\mathbf k_x,
\end{equation*}
where $\mathbf k_x=[k(x,x_1),\dots,k(x,x_n)]^{\top}$ and $K$ is the covariance matrix of the
observed inputs.  The kernel sets the correlation structure; we use the Mat\'ern-5/2 kernel, with
distance $r=\|x-x'\|$ and lengthscale $\ell$, giving $k(x,x')=\bigl(1+\sqrt5\,r/\ell+5r^2/(3\ell^2)\bigr)\exp\bigl(-\sqrt5\,r/\ell\bigr)$.  

Every query maximizes the acquisition over the candidate set (the grid of options for each task).  

The three standard acquisitions used in this paper
are
\begin{itemize}
  \item
$
\mathrm{EI}(x)=\bigl(\mu(x)-\hat y\bigr)\,\Phi\bigl(z(x)\bigr)+\sigma(x)\,\phi\bigl(z(x)\bigr),
\citep{MOCKUS1975428, jones1998efficient},
\quad
$
\item
$
\mathrm{PI}(x)=\Phi\bigl(z(x)\bigr),\quad
\citep{kushner_new_1964}
$
\item
$
\mathrm{UCB}(x)=\mu(x)+\sqrt{\beta}\,\sigma(x),
\citep{DBLP:conf/icml/SrinivasKKS10}
$
\end{itemize}
where $z(x)=(\mu(x)-\hat y)/\sigma(x)$,  with $\Phi$ and $\phi$ the standard normal CDF and PDF and $\beta$ the confidence parameter of
GP-UCB \citep{DBLP:conf/icml/SrinivasKKS10}.   \hapbo{} leaves every element of this loop unchanged except the final ranking, which it
multiplies by the bounded factor of \eqref{eq:hapbo}.

\section{The Prior Specification Language (PSL)}
\label{app:psl}

\begin{figure}[h]
  \centering
  \includegraphics[width=\linewidth]{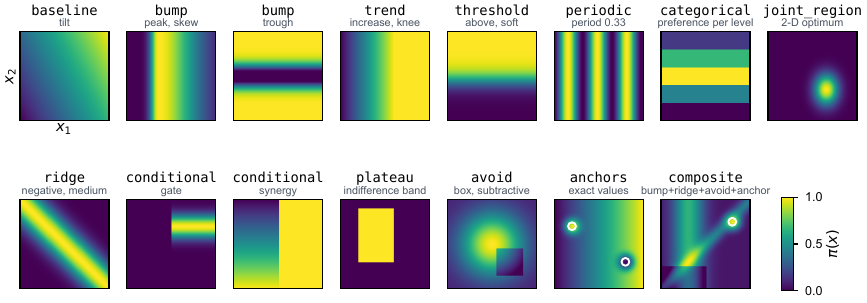}
  \caption{\textbf{PSL components as belief functions.}  Each panel is a minimal PSL document on
  a two-dimensional unit square ($x_1$ horizontal, $x_2$ vertical), compiled by
  Algorithm~\ref{alg:compile} into $\pi(x)\in[0,1]$.  In the \texttt{categorical} panel $x_2$ has
  five unordered levels; in the \texttt{anchors} panel the anchor values override a tilted baseline
  at the circled points.  The last panel stacks several components in one document.  Meta tokens
  (\texttt{smoothness}, \texttt{noise}, \texttt{modality}) and the \texttt{trust} block describe
  the objective rather than the optimum's location, so they have no panel.}
  \label{fig:psl_components}
  \end{figure}

A belief is one JSON document with three parts: 
\begin{itemize}
  \item a \texttt{schema} (the typed search space restated
by the source), 
\item a list of \texttt{beliefs}, 
\item\ a \texttt{trust} block.  
\end{itemize}
Table~\ref{tab:psl}
declares the vocabulary (PSL v1.0, Fig.~\ref{fig:psl_components}).  Three primitives (\texttt{bump}, \texttt{joint\_region},
\texttt{anchors}) form a complete basis, exact on finite candidate sets and dense in continuous
landscapes, with an explicit anchor-budget rate (App.~\ref{app:theory}).
The remaining tokens
are macros that name recurring patterns one token each; adding macros cannot reduce expressivity
(App.~\ref{app:theory}).  Every shaping belief carries a weight
$w\in[0,1]$ and a provenance tag; \texttt{anchors} carries no weight (its values are exact by
definition), and the meta tokens describe the objective rather than the optimum's location.

\begin{table}[h]
\centering
\caption{\textbf{Declaration of the PSL vocabulary.}  PSL v1.0: Required fields in parentheses; all shaping
tokens additionally require \texttt{weight} $\in[0,1]$ and \texttt{provenance} $\in$
\{\texttt{user\_stated}, \texttt{user\_confirmed}, \texttt{literature}, \texttt{llm\_domain},
\texttt{llm\_default}\}.  CI fields are 90\% intervals (half-width $=1.645\sigma$).}
\label{tab:psl}
\small
\begin{tabular}{@{}lp{4.55cm}p{5.6cm}@{}}
\toprule
PSL Token & Fields & Semantics \\
\midrule
\multicolumn{3}{@{}l}{\emph{Base layer}}\\
\texttt{baseline} & \texttt{level} $[0,1]$ (default $0.1$); \texttt{tilt} per-dim $[-1,1]$ &
default plausibility floor plus optional global slope \\
\midrule
\multicolumn{3}{@{}l}{\emph{Marginal shapes (one dimension)}}\\
\texttt{bump} & (\texttt{dim}, \texttt{mode}); \texttt{ci}; \texttt{skew} $[-1,1]$;
\texttt{sign} \{peak, trough\} & single peak or trough; split-Gaussian, $\sigma$ from the CI \\
\texttt{trend} & (\texttt{dim}, \texttt{direction} \{increase, decrease\}); \texttt{knee} &
monotone lean, optional saturation at the knee \\
\texttt{threshold} & (\texttt{dim}, \texttt{at}, \texttt{active} \{above, below\});
\texttt{sharpness} \{soft, sharp\} & regime change at a boundary (logistic step) \\
\texttt{periodic} & (\texttt{dim}, \texttt{period}); \texttt{phase} & oscillatory structure
(raised cosine) \\
\texttt{categorical} & (\texttt{dim}, \texttt{preferences} level$\to[0,1]$) & multiplicative
preference gate over unordered levels \\
\midrule
\multicolumn{3}{@{}l}{\emph{Interactions (two dimensions)}}\\
\texttt{joint\_region} & (\texttt{dims}, \texttt{mode} per-dim); \texttt{ci} per-dim & localized
joint optimum (product of per-dim Gaussians) \\
\texttt{ridge} & (\texttt{dims}, \texttt{relation} \{positive, negative\});
\texttt{width} \{narrow, medium, wide\}; \texttt{through} & trade-off line through two dimensions \\
\texttt{conditional} & (\texttt{when} \{dim, region\}, \texttt{then} nested belief);
\texttt{mode} \{gate, synergy\} & one dimension's shape activates within a region of another \\
\midrule
\multicolumn{3}{@{}l}{\emph{Regions and exact knowledge}}\\
\texttt{plateau} & (\texttt{region} per-dim interval) & indifference band (flat raise) \\
\texttt{avoid} & (\texttt{region} box \emph{or} \texttt{dim}/\texttt{mode}/\texttt{ci}) &
soft penalized region (subtractive; can never remove a candidate) \\
\texttt{anchors} & (\texttt{points} [\{coords, value $[0,1]$\}]); \texttt{bandwidth}
($>0$, default $0.15$) & exact values at known points, applied last \\
\midrule
\multicolumn{3}{@{}l}{\emph{Meta tokens (objective properties, not location beliefs)}}\\
\texttt{smoothness} & \texttt{level} / \texttt{per\_dim} \{very\_smooth\,\dots\,very\_rough\} &
suggests GP lengthscale \\
\texttt{noise} & (\texttt{level} \{low\,\dots\,very\_high\}); \texttt{repeats\_recommended} &
suggests GP noise floor \\
\texttt{modality} & \texttt{expectation} \{unimodal, few, many\}; \texttt{n\_optima} & suggests
exploration appetite \\
\midrule
\multicolumn{3}{@{}l}{\emph{Trust block (document level)}}\\
\texttt{trust} & (\texttt{overall\_confidence} \{very\_low, low, medium, high, very\_high\});
\texttt{notes} & source's stated confidence \\
\bottomrule
\end{tabular}
\end{table}

\textbf{Compilation of $\pi$.}  In plain terms, the compiler draws every stated shape on the
encoded unit cube, adds them up (penalties subtract), keeps the total nonnegative, applies the
preference gates, rescales the result to $[0,1]$, and finally lets exact anchors override the
field wherever the source knew the answer.  Algorithm~\ref{alg:compile} gives the procedure; each
$\mathrm{shape}_b$ and $\mathrm{gate}_g$ is the closed form declared in Table~\ref{tab:psl}. 

The normalization extrema are computed over a fixed quasirandom reference set
augmented with the anchors and every belief's declared mode or centre, so a narrow joint peak cannot be missed
by the normalizer.  The anchor blend uses compactly supported (Wendland-type) weights with default
bandwidth $0.15$ ($\psi_j$ vanishes beyond distance $\text{bandwidth}\cdot\sqrt d$ from $z_j$, as distances are normalized by $\sqrt d$) and is exact at anchored points, which is what makes the anchor basis exact and
dense (App.~\ref{app:theory}).  Compilation is deterministic.

The compiler draws every stated shape on the encoded unit cube
and combines them in five steps.  Let $b_0$ and $\tau$ be the baseline level and tilt, and $w_b$
the weight of belief $b$.

\emph{(i) Encoding.}  Each of the $d$ dimensions is mapped onto $[0,1]$ (log-spaced ranges on the
  log scale, ordinal and categorical levels by rank), so that a configuration becomes $x\in[0,1]^d$.

  \emph{(ii) Shapes.}  Each belief $b$ is drawn as its closed form $\mathrm{shape}_b(x)\in[0,1]$
  (Table~\ref{tab:psl}).
  
\emph{(iii) Sum, clamp and gates.}   The shapes are added, each scaled by its weight; penalties
  (\texttt{avoid} and troughs) subtract:
  \begin{equation}
    s(x) \;=\; b_0 + \tau^\top\!\big(x-\tfrac12\big) \;+\; \sum_{b} \pm\, w_b\,\mathrm{shape}_b(x).
    \label{eq:score}
  \end{equation}
The score is clamped at zero and multiplied by the gates
(categorical preferences and synergy conditionals):
\begin{equation}
  \tilde s(x) \;=\; \max\{s(x),0\}\,\prod_{g}\mathrm{gate}_g(x),
  \qquad \mathrm{gate}_g(x)\in[0,1].
  \label{eq:gates}
\end{equation}

\emph{(iv) Normalization.}  The score is rescaled with its extrema $(\ell,u)$ over the fixed
reference set $R$:
\begin{equation}
  \pi_0(x) \;=\; \mathrm{clip}\!\left(\frac{\tilde s(x)-\ell}{u-\ell},\,0,\,1\right).
  \label{eq:norm}
\end{equation}

\emph{(v) Anchors.}  Anchor points $(z_j,v_j)$ override the field locally through compactly
supported weights $\psi_j$:
\begin{equation}
  \pi(x) \;=\; \Big(1-\sum_j\psi_j(x)\Big)\,\pi_0(x) + \sum_j \psi_j(x)\,v_j,
  \label{eq:anchors}
\end{equation}
so that $\pi(z_j)=v_j$ exactly.  Algorithm~\ref{alg:compile} gives the same procedure; each
$\mathrm{shape}_b$ and $\mathrm{gate}_g$ is the closed form declared in Table~\ref{tab:psl}.

\begin{figure}[h]
\centering
\includegraphics[width=\linewidth]{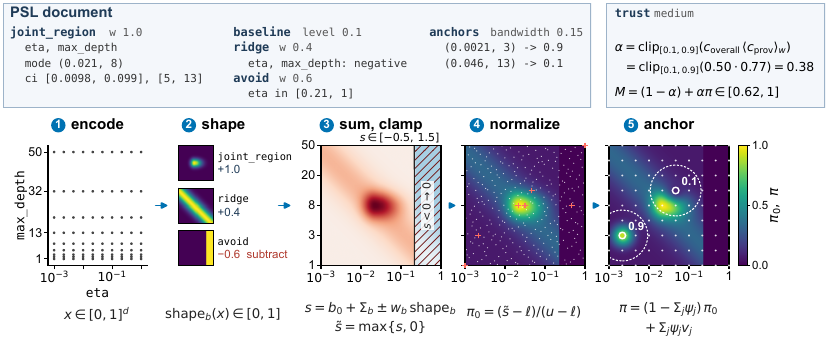}
\caption{\textbf{Compiling one PSL document.} The sample PSL document (top) refers to the XGBoost task on \texttt{kc1} of App.~\ref{app:psl}. It lives on the  space of two XGBoost parameters:
\texttt{eta}$\times$\texttt{max\_depth}. \textbf{Step 1}~Irregular document candidates (dots) become a regular grid.
\textbf{Step 2}~Closed forms are applied. \textbf{Step 3}~A signed sum is performed, and clamped at zero. \textbf{Step 4}~Rescaling on $R$ (dots: Halton points; crosses: declared points);
 $u=1.5$ is attained at the declared mode. \textbf{Step 5}~Anchors (circles: $z_j$, labelled with $v_j$) will override the field within radius
$0.15\sqrt d$ (dashed: support of $\psi_j$). Top right: $\alpha$ from the card to Eq.~\ref{eq:alpha}.}
\label{fig:compile-trace}
\end{figure}

\begin{algorithm}[h]
\caption{Compilation of the belief function $\pi$ from a validated PSL document.}
\label{alg:compile}
\begin{algorithmic}[1]
\Require validated document $P=(\texttt{schema},\texttt{beliefs},\texttt{trust})$; query points
$x$ on the encoded unit cube $[0,1]^d$
\Ensure belief function $\pi:\mathcal X\to[0,1]$
\State $s(x)\gets\texttt{baseline.level}+\texttt{tilt}^{\top}\!\bigl(x-\tfrac12\bigr)$
\Comment{base layer}
\ForAll{additive beliefs $b\in\{$\texttt{bump}, \texttt{trend}, \texttt{threshold},
\texttt{periodic}, \texttt{joint\_region}, \texttt{ridge}, \texttt{plateau}, \texttt{avoid}$\}$}
  \State $s(x)\gets s(x)\pm w_b\,\mathrm{shape}_b(x)$
  \Comment{\texttt{avoid} and troughs subtract}
\EndFor
\State $s(x)\gets\max\{s(x),\,0\}$
\Comment{clamp}
\ForAll{gates $g\in\{$\texttt{categorical}, synergy \texttt{conditional}$\}$}
  \State $s(x)\gets s(x)\cdot\mathrm{gate}_g(x)$
  \Comment{gates take values in $[0,1]$}
\EndFor
\State $R\gets8{,}192$ unscrambled Halton points $\,\cup\,$ \{declared modes, anchor coord,
corners, center, \dots\}
\Comment{reference set}
\State $(\ell,u)\gets\bigl(\min_{r\in R}s(r),\;\max_{r\in R}s(r)\bigr)$
\State $\pi(x)\gets\operatorname{clip}\bigl((s(x)-\ell)/(u-\ell),\,0,\,1\bigr)$
\Comment{min--max normalization}
\ForAll{anchor sets with points $(z_j,v_j)$ and weights $\psi_j$}
  \State $\pi(x)\gets\bigl(1-\textstyle\sum_j\psi_j(x)\bigr)\,\pi(x)+\sum_j\psi_j(x)\,v_j$
  \Comment{applied last; $\pi(z_j)=v_j$ exactly}
\EndFor
\State \Return $\pi$
\end{algorithmic}
\end{algorithm}

\textbf{Compilation of $\alpha$} (Eq.~\ref{eq:alpha} of the main text). $\alpha$ represents the overall confidence assigned to $\pi$. 
Choosing the most appropriate $\alpha$ for a given task can be a standalone research question. The current research is designed in a modular way, and we decided to opt for a simplification for the choice of assigning $\alpha$.
In this paper, the overall confidence maps to $c_{\text{overall}}\in\{0.20,0.35,0.50,0.70,0.90\}$ for the five levels; each belief's provenance
maps to $c_{\text{provenance}}\in\{1.00,0.90,0.85,0.65,0.40\}$ for:
\begin{itemize}
  \item \texttt{user\_stated} (highest confidence),
  \item \texttt{user\_confirmed},
  \item \texttt{literature},
  \item \texttt{llm\_domain},
  \item \texttt{llm\_default} (lowest confidence).
\end{itemize}

$\alpha=\operatorname{clip}(c_{\text{overall}}\cdot\langle c_{\text{provenance}}\rangle_w,\,
0.1,\,0.9)$ with the average weighted by the belief weights. In practice, this leads to a medium-confidence document built
entirely on model-internal knowledge, compiling to $\alpha=0.325$ in most of the present work; only a user-stated,
very-high-confidence document raises this value (e.g., the cortical priors sourced by a human expert).

Several different choices are possible and the \hapbo{} method is designed to accommodate a future $\alpha$ selection module.

The original $\alpha$-$\pi$BO work \citep{GUAYHOTTIN2025113039} suggested a default $\alpha=0.5$ for all priors, which would be a reasonable choice here that would not affect the results qualitatively.  The present work uses the more  nuanced $\alpha$ of Eq.~\ref{eq:alpha} to illustrate that the PSL expressivity can be modulated by design, by specific choice rules (our implementation), or even by LLM feedback.

\textbf{Meta tokens and the surrogate.}  \texttt{smoothness}, \texttt{noise}, and
\texttt{modality} can configure the GP (per-dimension lengthscales, noise floor, exploration).
For a fixed surrogate, the compiled field $\pi$ enters the noiseless EI guarantee through a
multiplicative constant.  The separate noisy GP-UCB bound also contains an additive
misplaced-trust term, which can grow linearly under static trust (App.~\ref{app:theory}).  In all
campaigns reported in this paper the meta-token channel is disabled. Future work can investigate the opportunity to further guide the BO approach by having the LLM condition the GP properties. 
The authors have considered this capacity to be beyond the scope of the current research. 

\textbf{A real elicited document.}  The prior compiled for XGBoost on the \texttt{kc1} task, as
returned by the elicitor (numeric values rounded for display):

\begin{Verbatim}[frame=single,fontsize=\footnotesize]
{ "schema": { "objective": "maximize",
              "dimensions": [ {"name": "colsample_bytree", ...},
                              {"name": "eta", ...},
                              {"name": "max_depth", ...},
                              {"name": "reg_lambda", ...} ] },
  "beliefs": [
    { "token": "baseline", "level": 0.1 },
    { "token": "joint_region", "dims": ["eta", "max_depth"],
      "mode": { "eta": 0.021, "max_depth": 8 },
      "ci":   { "eta": [0.0098, 0.099], "max_depth": [5, 13] },
      "weight": 1.0, "provenance": "llm_domain",
      "note": "Standard sweet spot for XGBoost: moderate learning
               rate (~0.01-0.1) with moderate depth (5-10) balances
               bias/variance effectively on tabular data." },
    { "token": "bump", "dim": "colsample_bytree", "mode": 0.6,
      "ci": [0.4, 0.8], "weight": 0.8, "provenance": "llm_domain",
      "note": "Feature subsampling around 0.5-0.7 typically reduces
               overfitting without losing too much signal." },
    { "token": "bump", "dim": "reg_lambda", "mode": 0.46,
      "ci": [0.099, 10.1], "weight": 0.6, "provenance": "llm_domain",
      "note": "Moderate L2 regularization helps stability on
               imbalanced datasets; very high values underfit." },
    { "token": "smoothness", "level": "medium" },
    { "token": "noise", "level": "medium", "repeats_recommended": false },
    { "token": "modality", "expectation": "few" } ],
  "trust": { "overall_confidence": "high",
             "notes": "Based on extensive empirical literature regarding
                       XGBoost hyperparameter tuning on tabular
                       classification tasks." } }
\end{Verbatim}

\section{Task cards and elicitation protocol}
\label{app:cards}

\textbf{Card structure.}  Every source receives the same three-part card: a domain description, the
typed search space, and the optimization description (objective, direction, budget, known
constraints), preceded by a fixed instruction to express the belief only in the closed PSL
vocabulary.  Cards are de-identified, they do not name the public benchmark or
dataset (the domain is described only by model class and task type), and they carry no
hyperparameter guidance or expectations about the response surface. The elicitor must place the
prior from its own knowledge.  The card for the XGBoost defect-prediction task (parameter value
lists abridged):

\begin{quote}\small
\emph{Domain.}  System under study: hyperparameter optimization of an XGBoost
gradient-boosted-tree classifier on a software-defect-prediction task (about 2,000
records, about 20 numeric features, 2 strongly imbalanced classes).  The four knobs
are the learning rate \texttt{eta} (log), \texttt{max\_depth} (log), \texttt{colsample\_bytree}
(per-tree feature-subsampling fraction), and the L2 leaf regularization \texttt{reg\_lambda}
(log).  General gradient-boosted-tree
hyperparameter-tuning knowledge applies.

\emph{Search space.}  \texttt{colsample\_bytree}: ordinal, 10 linear values in $[0.1,1.0]$;
\texttt{eta}: ordinal, 10 log-spaced values in $[2^{-10},1]$; \texttt{max\_depth}: ordinal,
$\{1,2,3,5,8,13,20,32,50\}$; \texttt{reg\_lambda}: ordinal, 10 log-spaced values in
$[2^{-10},1024]$.

\emph{Optimization description.}  Objective: mean validation accuracy (fraction correct) on a held-out split, averaged over several random
seeds, higher is better.  Direction: maximize.  Evaluation budget: about 8 to 32 evaluations
(low-budget regime).  Known constraints or prior observations: none.

\end{quote}

\textbf{Human expertise enters as text on the same card.}  For the cortical mixed
(LLM$+$human) prior, the card's third section embeds the expert's statement verbatim:

\begin{quote}\small
A domain expert who has run similar experiments on this preparation states: ``Usually the top row
is the most effective, and the two outer columns of the array are the most effective.''  
\end{quote}

\noindent The provenance tags of the resulting beliefs (\texttt{user\_stated} vs.\
\texttt{llm\_domain}) record which parts came from the human and set the compiled trust through
Eq.~\ref{eq:alpha}; there is no separate numeric merge step. 

For the \emph{human-only prior}, the elicitor is given access to a summary of data from 17 subjects that are not contained in the dataset (as in \citep{bonizzato2023autonomous}), transcribing expert maps
directly.

\textbf{Adversarial priors.}  The matched adversarial condition replaces the compiled belief with a
programmatically inverted one, which was selected to be a \texttt{joint\_region} bump placed at the task's empirical
anti-optimum (the worst grid configuration), same weight and provenance as the faithful prior. So
the adversarial arm consumes exactly the same interface with exactly misplaced content.  Baselines
that consume text instead of a compiled prior receive a hostile task description through
their native interface.

\textbf{Elicitation protocol.}  The priors are elicited from a locally served open-weights
model (Qwen3.5-122B) at temperature $0$, one prior per task (synthetic panel: three draws per card at temperature $0.6$;
  elicitor comparison:Appendix~\ref{app:ladder}), offline, before any optimization.  The
first attempt uses schema-guided decoding against the PSL grammar; if validation or compilation
fails, the exact validator error is fed back for up to four corrective turns (typical cost: one to
two calls, ${\sim}15{,}000$ tokens total per HPO task).  

\section{Experimental details and cost}
\label{app:details}

\textbf{Surrogate and acquisition.}  All internal arms (\hapbo{}, its ablation variants, vanilla
BO) share a Mat\'ern-5/2 GP (Appendix~\ref{app:variants} also tests RBF and Mat\'ern-1/2 kernels).  The acquisition is EI with $\xi=0$ , except in Appendix~\ref{app:variants}, which tests EI/PI/UCB; 
already-queried candidates are excluded from selection in the noiseless benchmarks.
The single $n_{\text{initial}}=1$ random point per seed is identical across \hapbo{}, vanilla, and random search.
Seeds 1001--1010.  Prior warm-starting is disabled everywhere except the injection-site ablation arm that studies it and $\pi$-BO, whose first initial point is the prior mode as in \citet{hvarfner2022piboaugmentingacquisitionfunctions} (App.~\ref{app:ablation-tables}).

\textbf{Estimands and normalizations.}  HPOBench: simple regret against the tabulated grid optimum;
figure curves normalize per task by the grid span (optimum minus mean grid score); the adversarial
figure normalizes by vanilla's mean first-query regret.  Cortical: ground truth response of the electrode currently considered best divided by the optimum one, averaged over ten seeds; comparisons use all 40 rat--muscle maps, with mean $\pm$ SEM over the maps (technical replicates); budget 12 queries.  Synthetic: per-function regret normalized by vanilla's mean first-query regret and aligned
by fraction of each function's budget (Hartmann-6: 40, Styblinski--Tang-7: 50, Levy-10: 50,
Ackley-15: 150 queries).  SOLAR: best feasible efficiency, zero until the first feasible design.

\textbf{Synthetic prior fidelity.}  For each analytic function the elicitor receives a de-named
card carrying a perturbed closed-form model of the objective whose correlation with the truth is
controlled ($\rho\approx0.9$, $0.5$, $0.1$), and transcribes it into PSL under a maximum-fidelity
instruction set.  Prior quality is therefore an experimental variable; the compiled trust of Eq.~\ref{eq:alpha} is used unchanged.

\textbf{Detail of prior elicitation quality in the synthetic experiment.}  Table~\ref{tab:synth-prior-quality}
shows, for each function and card fidelity, how closely the compiled prior $\pi$ follows the card's model $M$ and the
true objective $f$, and the value of $\pi$ at the global optima of each ($x^\star_M$, $x^\star_f$).  At every
fidelity the elicited prior has a peak aligned with the card model that elicited it.  The card model's optimizer does not coincide with the true one.  The global agreement $\rho(\pi,M)$ stays well below one with the Qwen3.5-122B elicitor used throughout;
see Appendix~\ref{app:ladder} for how stronger elicitors transcribe the card model with more accuracy and detail, and improve global agreement.

\begin{table}[h]
\centering
\caption{\textbf{Prior elicitation quality in the synthetic experiment.}  $M$: the closed-form model printed on the
card; $f$: the true objective; $\pi$: the compiled PSL prior (Qwen3.5-122B; mean over the three elicitation draws).
$\rho(\cdot,\cdot)$: Pearson correlation over $20{,}000$ uniform points of the search box.  $\pi(x^\star_f)$, $\pi(x^\star_M)$: value of $\pi$ at the global
optimizer of $f$ and of $M$.  }
\begin{tabular}{lccccc}
\toprule
Function & $\rho(M,f)$ & $\rho(\pi,M)$ & $\rho(\pi,f)$ & $\pi(x^\star_f)$ & $\pi(x^\star_M)$\\
\midrule
\textbf{$\boldsymbol{\rho=0.9}$, all functions} & \textbf{0.90} & \textbf{0.38} & \textbf{0.41} & \textbf{0.50} & \textbf{0.90}\\
\quad Hartmann-6 & 0.91 & 0.41 & 0.39 & 0.08 & 0.72\\
\quad Styblinski--Tang-7 & 0.90 & 0.15 & 0.12 & 0.00 & 0.86\\
\quad Levy-10 & 0.90 & 0.41 & 0.51 & 0.91 & 1.00\\
\quad Ackley-15 & 0.90 & 0.55 & 0.63 & 1.00 & 1.00\\
\midrule
\textbf{$\boldsymbol{\rho=0.5}$, all functions} & \textbf{0.50} & \textbf{0.34} & \textbf{0.30} & \textbf{0.62} & \textbf{0.98}\\
\quad Hartmann-6 & 0.49 & 0.56 & 0.45 & 0.79 & 0.93\\
\quad Styblinski--Tang-7 & 0.51 & 0.21 & $-$0.05 & 0.01 & 0.99\\
\quad Levy-10 & 0.51 & 0.35 & 0.40 & 0.75 & 1.00\\
\quad Ackley-15 & 0.50 & 0.24 & 0.39 & 0.93 & 1.00\\
\midrule
\textbf{$\boldsymbol{\rho=0.1}$, all functions} & \textbf{0.11} & \textbf{0.15} & \textbf{0.08} & \textbf{0.14} & \textbf{1.00}\\
\quad Hartmann-6 & 0.12 & 0.03 & 0.08 & 0.14 & 1.00\\
\quad Styblinski--Tang-7 & 0.11 & 0.20 & $-$0.15 & 0.13 & 1.00\\
\quad Levy-10 & 0.11 & 0.33 & 0.23 & 0.11 & 1.00\\
\quad Ackley-15 & 0.11 & 0.06 & 0.16 & 0.20 & 1.00\\
\bottomrule
\end{tabular}

\label{tab:synth-prior-quality}
\end{table}

\textbf{SOLAR constraints.}  \hapbo{} fits one GP per constraint and multiplies the (shaped)
acquisition by the product of per-constraint feasibility probabilities $\Phi((0-\mu_i)/\sigma_i)$;
simulator crashes are modeled by an additional GP whose probability-of-success factor multiplies
the score, and crashed observations are excluded from the objective and constraint models.  The
incumbent is the best feasible, non-crashed observation.

\textbf{Cost accounting.}  Table~\ref{tab:cost} reports per-query wall-clock and token costs on
HPOBench.  Wall-clock is the median budget-16 run divided by budget on the archived hardware; the
LLAMBO and BORA token figures are measured on a 10-run subset rerun and LGBO's from run telemetry; hardware differs across some
archived external runs, so these are order-of-magnitude costs, not a controlled benchmark.  The
locally served model has no marginal provider fee; its compute and energy costs are nonzero.

\begin{table}[h]
\centering
\caption{\textbf{Per-query cost on HPOBench} (Qwen3.5-122B served locally), averaged across the
five families.  \emph{Setup tokens} is the one-off prior elicitation, compiled once per task;
\emph{tokens/query} is the in-loop LLM cost, zero when the model is consulted only offline.}
\begin{tabular}{lccc}
\toprule
Method & wall-clock/query (s) & setup tokens (once) & tokens/query\\
\midrule
HapBO & 0.20 & 15{,}001 & 0\\
ColaBO & 104.30 & 1{,}770 & 0\\
LLAMBO & 28.59 & -- & 79{,}650\\
LGBO & 12.44 & -- & 566\\
BORA & 4.08 & -- & 266\\
vanilla BO & 0.20 & -- & --\\
random & 0.20 & -- & --\\
\bottomrule
\end{tabular}

\label{tab:cost}
\end{table}

\section{Fixed-belief ablations: trajectories and endpoints}
\label{app:ablation-tables}

This appendix reports the two ablations summarized in Sec.~\ref{sec:ablations}.
Both hold the belief fixed, so that differences can be attributed to how a prior
is used rather than to what it says: every arm consumes one compiled prior
$(\pi,\alpha)$ per task and shares the same GP surrogate, base EI, and ten seeds; only the stated axis differs.

\begin{figure}[t]
\centering
\includegraphics[width=\linewidth]{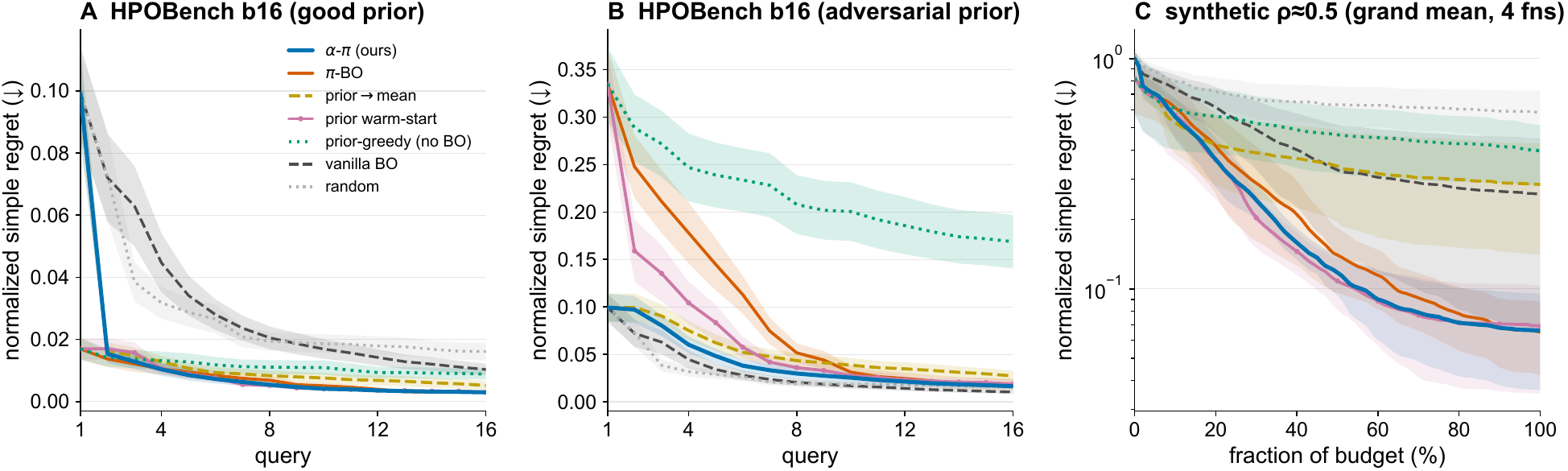}
\caption{\textbf{Prior injection site ablation.}  One compiled prior $(\pi,\alpha)$ per
task; every arm shares the GP surrogate, base EI, and curves are
query-wise mean $\pm$ SEM over the 40 HPOBench tasks (16 queries) and the four analytic functions aligned by fraction of budget. \textbf{A} faithful HPO prior, \textbf{B}  adversarial HPO prior, \textbf{C} synthetic $\rho\approx0.5$, grand mean over the four analytic functions.  
}
\label{fig:ablations}
\end{figure}

\textbf{Injection site} (Fig.~\ref{fig:ablations}).
With the representation fixed, the prior enters through the bounded acquisition
multiplier of \hapbo{}, the floor-free $\pi$-BO multiplier, the surrogate mean
(prior-to-mean), or the initial queries (prior-warm-start), bracketed by a
prior-only greedy arm and no-prior controls.
The $\pi$-BO arm follows \citet{hvarfner2022piboaugmentingacquisitionfunctions} as published ($\beta=N/10$
with $N$ the number of BO iterations). Its first initial point is the mode of the prior.

\begin{figure}[p]
\centering
\includegraphics[width=0.92\linewidth]{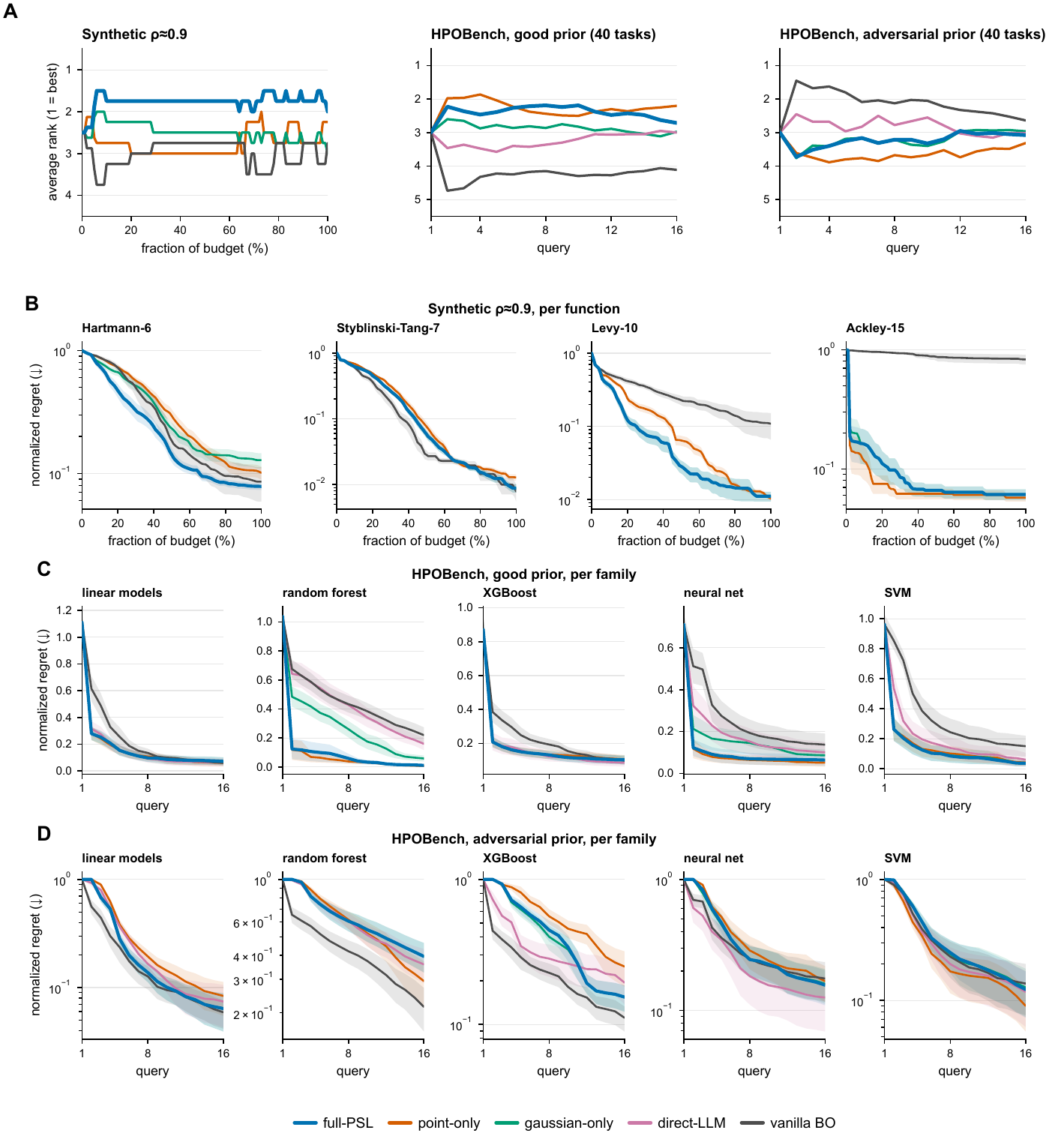} %
\caption{\textbf{Prior representation ablation.}  The injection site is fixed at the bounded $\alpha-\pi$BO
multiplier; only how the belief becomes $\pi$ varies: full PSL (ours), point-only, Gaussian-only, direct-LLM elicitation,
and vanilla BO (no prior).  Grand means in Fig.~\ref{fig:representation-main}.  \textbf{A:} average task-wise rank (1 = best) in the same three conditions (synthetic suite at $\rho\approx0.9$, aligned by fraction of budget; HPOBench with the good and with the adversarial prior).
\textbf{B--D:} per synthetic function (\textbf{B}), and per HPOBench model family with the good (\textbf{C}) and with the adversarial prior (\textbf{D}).
Direct-LLM is absent from the synthetic suite, because its fixed anchor set degenerates in 7--15D.  Ten seeds per task (synthetic prior arms: three documents $\times$ ten seeds per function); mean $\pm$ SEM across tasks; lower is better.}
\label{fig:representation}
\end{figure}

\textbf{Representation} (Figs.~\ref{fig:representation-main} and~\ref{fig:representation}).
With the site fixed at the bounded multiplier, only the language through which
the belief becomes $\pi$ varies: PSL against point-only, Gaussian-only, and
direct-LLM elicitation. PSL's richness pays off on trustworthy high-dimensional synthetic priors, in particular. Direct-LLM elicitation is reported on the
finite benchmarks only, since a fixed coarse anchor grid degenerates in 7--15 dimensions. We evaluate the five arms as in the main evaluation (grand mean, Fig.~\ref{fig:representation-main}; average
task-wise rank, Fig.~\ref{fig:representation}A; per-function/family, Fig.~\ref{fig:representation}B--D) on the synthetic suite at $\rho\approx0.9$ and on the 40 HPOBench
tasks with the good and adversarial prior. Full PSL wins on the synthetic suite. On HPOBench with the good prior, full PSL is aligned with  point-only and above the other options. 
With the inverted-map prior, full PSL performs better than point-only, and similarly to the other options. 

\begin{figure}[t]
\centering
\includegraphics[width=\linewidth]{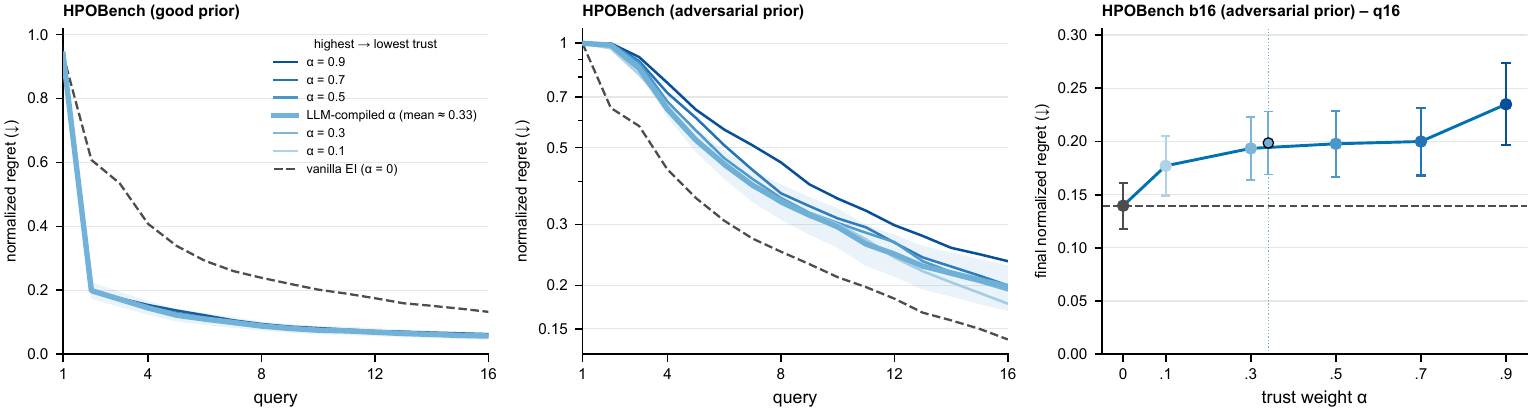}
\caption{\textbf{Trust weight $\alpha$ of \hapbo{}.}  \hapbo{} ($\alpha$-$\pi$) with the trust
weight fixed at $\alpha=0.9$, $0.7$, $0.5$, $0.3$, or $0.1$ instead of the LLM-compiled $\alpha$; 40 HPOBench tasks $\times$ 10 seeds, 16 queries, matched random start; mean $\pm$ SEM over tasks.
\textbf{Left:} good prior.  \textbf{Middle:} misleading prior.  \textbf{Right:} misleading prior, final regret against
$\alpha$.  Colors run from the highest (darkest) to the lowest trust; dashed: vanilla BO ($\alpha=0$); LLM-compiled
$\alpha$: thick line (left, middle) and outlined marker (right).}
\label{fig:alpha-sweep}
\end{figure}

\textbf{Trust weight} (Fig.~\ref{fig:alpha-sweep}).
How important is the confidence stated by the LLM? It enters \hapbo{} through the trust weight $\alpha$
(\eqref{eq:alpha}). We keep each task's compiled $\pi$ and replace the compiled $\alpha$ by a fixed
$\alpha\in\{0.9,0.7,0.5,0.3,0.1\}$ on the 40 HPOBench tasks (ten seeds, 16 queries). With the good prior (task card), every $\alpha$ in $[0.1,0.9]$ gives about 55\% lower
final regret than vanilla EI (grid-span normalized regret at query 16: vanilla $0.133$, $\alpha=0.9$ $0.063$), and
$\alpha\in[0.1,0.5]$ is indistinguishable from the compiled $\alpha$. With the misleading prior, regret at query 16 rises with $\alpha$. 
The trust weight thus bounds the loss from a misleading prior. In different optimization problems, regret might decrease with increasing $\alpha$ for good priors~\citep{GUAYHOTTIN2025113039}

\begin{figure}[t]
\centering
\includegraphics[width=\linewidth]{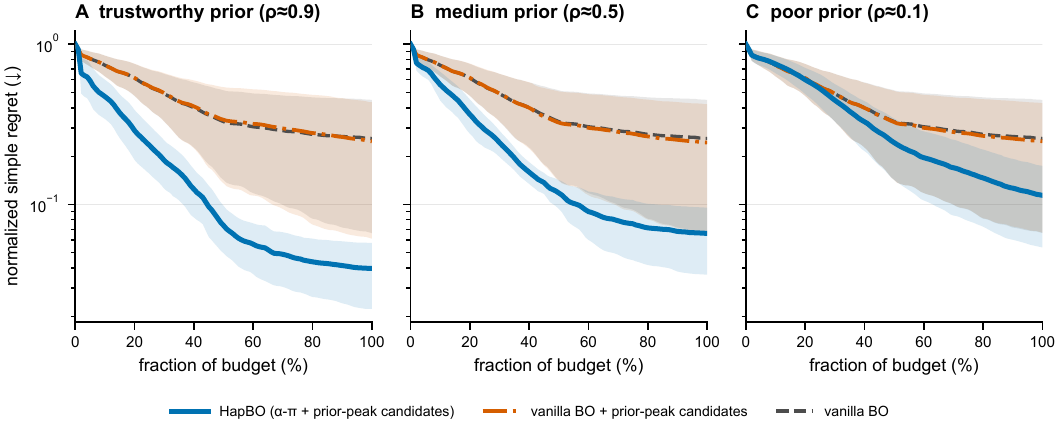}
\caption{\textbf{Prior-peak candidates alone do not transmit the prior (synthetic panel).}
Normalized simple regret as in Fig.~\ref{fig:synthetic} under trustworthy (A), medium (B), and poor (C) priors.
\hapbo{} against vanilla BO given
the same prior-maximizer candidates with an unweighted acquisition, and vanilla BO.}
\label{fig:peaks-control}
\end{figure}

\textbf{Note on optimization grids in continuous domains} (Fig.~\ref{fig:peaks-control}).
On the finite benchmarks the acquisition is maximized over the whole candidate grid, so the multiplier
scores every option, the prior's maximizers included. A continuous domain has no grid: each query
maximizes the acquisition from a finite set of candidate points (2048 scrambled Sobol points, the best
eight refined by L-BFGS-B). In 7--15 dimensions such a set almost never falls inside a narrow prior mode,
so on the synthetic benchmarks (Fig.~\ref{fig:synthetic} and Fig.~\ref{fig:representation}) every prior arm
also receives the maximizers of its own $\pi$ as candidates.
Including the prior's peaks among the candidate points is the obvious choice (since they're likely candidates for final acquisition function argmax), but to prove that it does not constitute an additional mechanism
by which the prior is transmitted, we show that vanilla BO given the same candidates does not improve (Fig. \ref{fig:peaks-control}).

\section{Prior sources: compiled fields, example maps, and trajectories}
\label{app:sources-full}

\begin{figure}[h]
\centering
\includegraphics[width=0.72\linewidth]{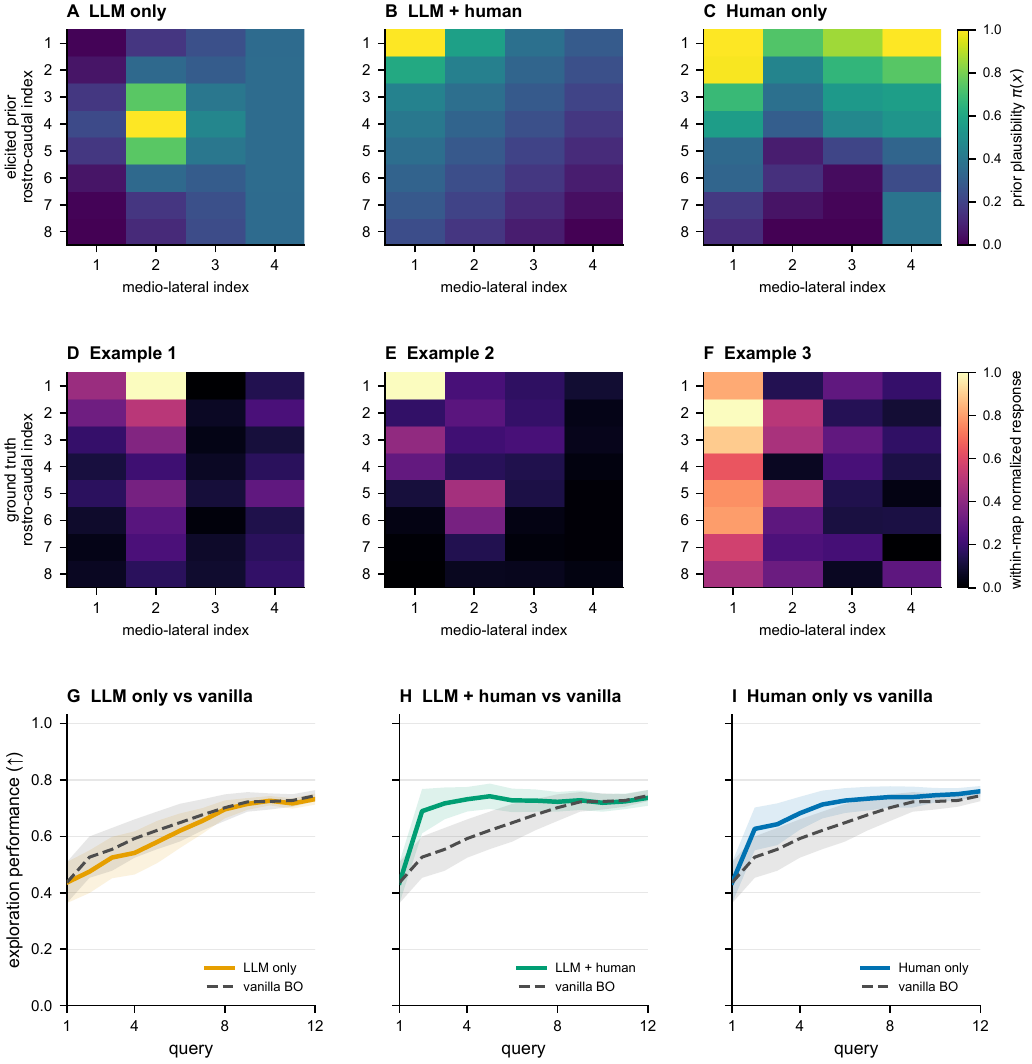}
\caption{\textbf{Prior source and cortical optimization, complete panel.}  \textbf{A--C:} compiled
belief functions for LLM-only, LLM$+$human, and human-only knowledge.  \textbf{D--F:} three
example ground-truth response maps from the NeuroMOSAICS dataset, each normalized independently to
expose spatial structure; they are not paired with the priors above.  \textbf{G--I:} each prior
mode against matched vanilla BO over twelve queries (budget 12); mean $\pm$ SEM clustered over six rats after
averaging muscles and ten seeds.}
\label{fig:sources-full}
\end{figure}

This appendix reports the prior-source comparison on the NeuroMOSAICS Cortical panel,
together with the objects the trajectories consume.  The three compiled
belief functions (Fig.~\ref{fig:sources-full}A--C) show how differently the same
interface renders the three knowledge sources: the LLM-only prior encodes a smooth
somatotopic gradient, the LLM$+$human prior transcribes the expert's rostral-row and
edge-column preferences, and the human-only prior reconciles data from previous
subjects (rats tested under the same conditions but not included in the dataset).
The example ground-truth response maps (Fig.~\ref{fig:sources-full}D--F) illustrate the spatial structure and
cross-subject variability of the underlying objective; they are shown for context
and are not paired with the priors above them.   Fig.~\ref{fig:sources-full}G--I shows each prior mode
against its matched vanilla BO trajectory.

  \section{HPOBench: per-family trajectories}

  \label{app:hpo-family}
  
\begin{figure}[h]
  \centering
  \includegraphics[width=\linewidth]{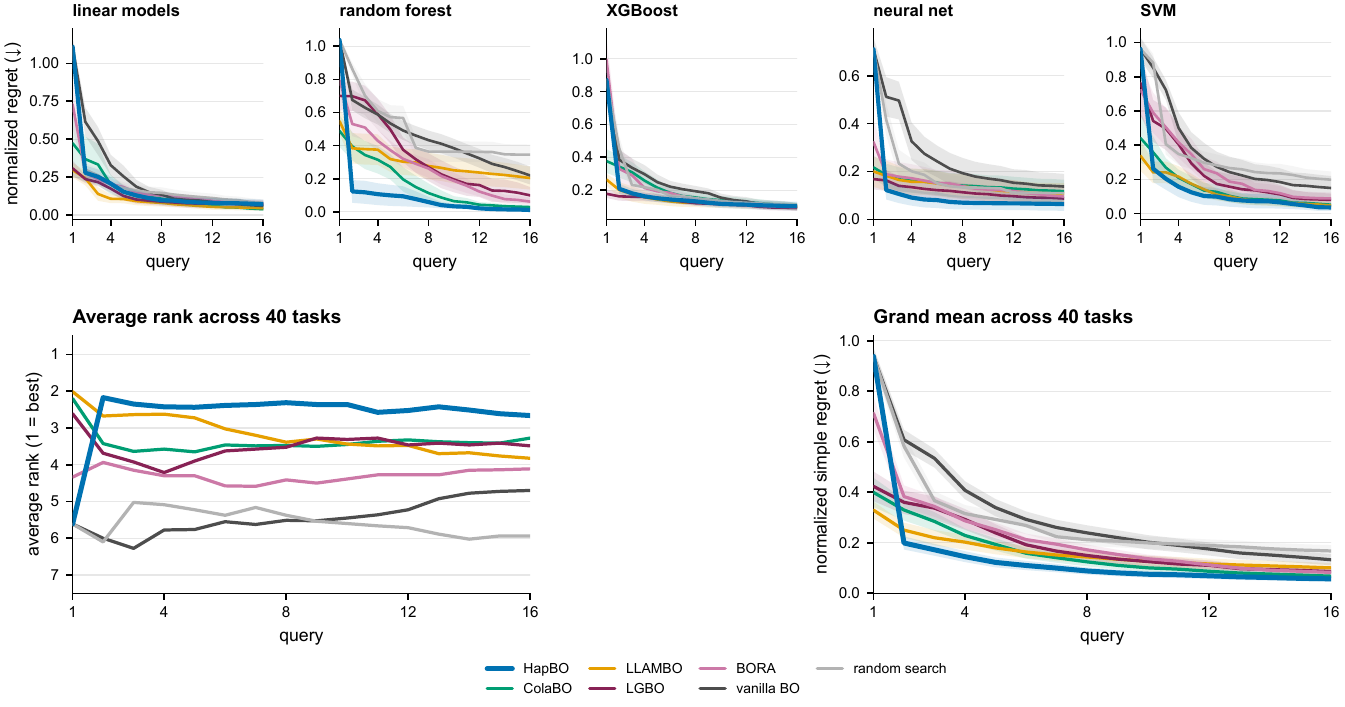}
  \caption{\textbf{HPOBench, per model family.}  Normalized simple regret for the five families (top),
  average task-wise rank and grand mean over 40 tasks (bottom), using matched \hapbo{} and every
  method's native budget-16 curve.  Regret is normalized per task by the full-grid span (optimum minus
  mean grid score); bands are $\pm$SEM across tasks after averaging ten seeds within each task.
  Every method completes all 400 task--seed runs.  The family panels show the aggregate gain of the
  main text is not driven by one model class.}
  \label{fig:hpo-family}
  \end{figure}

  Figure~\ref{fig:hpo-family} expands the compact comparison of the main text (Fig.~\ref{fig:hpo})
  into its five per-family panels.  Matched \hapbo{} consumes the compiled HPO prior from the same
  random start as its controls, and every method shows its own curve. 

  \label{app:adv-family}

\begin{figure}[h]
\centering
\includegraphics[width=\linewidth]{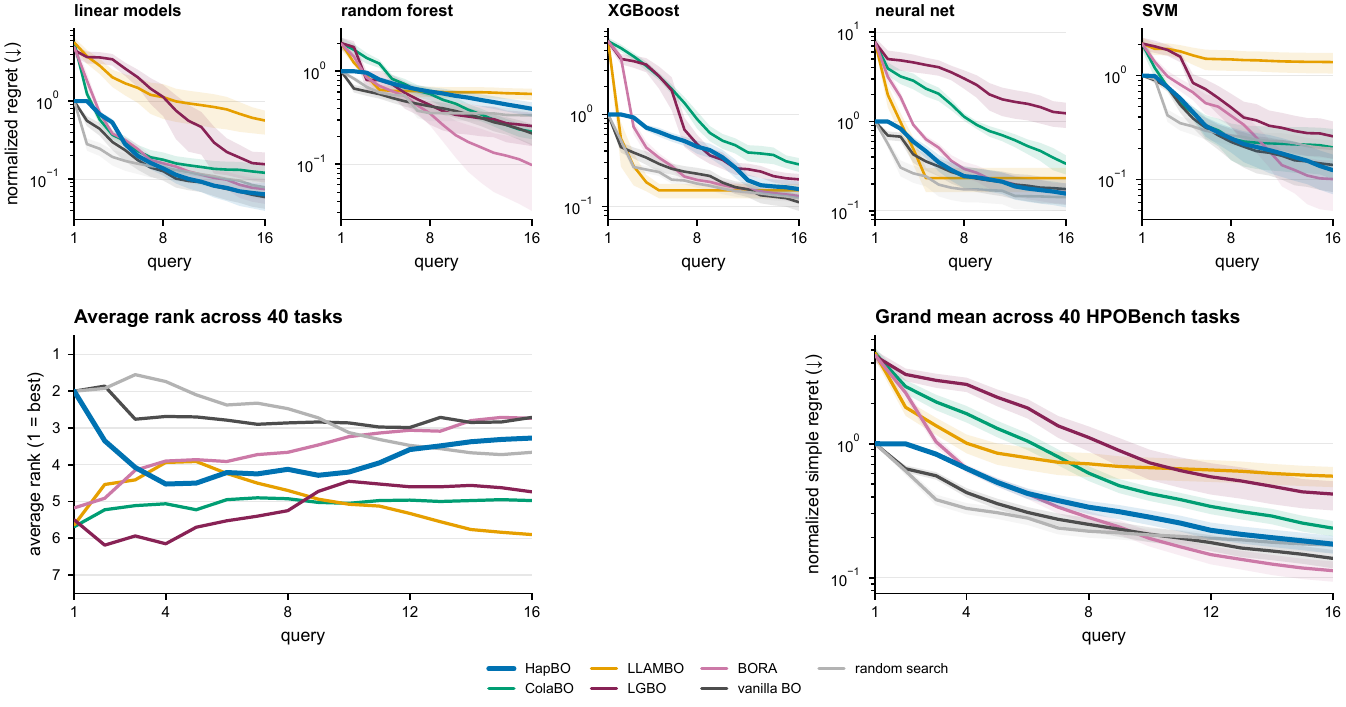}
\caption{\textbf{Adversarial HPOBench, per model family.}  Normalized simple regret for the five
families (top), average task-wise rank and grand mean over 40 tasks (bottom), using matched
\hapbo{} and the same hostile description through each method's native interface.  The rank panel
uses the task grid complete for every method; regret is normalized by vanilla's mean first-query
regret; bands are $\pm$SEM across tasks after averaging ten seeds.  BORA completes 400/400, LLAMBO
400/400, and ColaBO 400/400 task--seed endpoints.}  
\label{fig:adv-family}
\end{figure}

Figure~\ref{fig:adv-family} expands the compact adversarial comparison of the main text
(Fig.~\ref{fig:robust}) into its five per-family panels.  Each method receives the same hostile
task description through its native interface, and matched \hapbo{} consumes the compiled
adversarial prior (a bump at each task's empirical anti-optimum) from the same random start as
its controls. 

\section{\hapbo{} across BO variants}
\label{app:variants}

\begin{figure}[htbp]
\centering
\includegraphics[width=\linewidth]{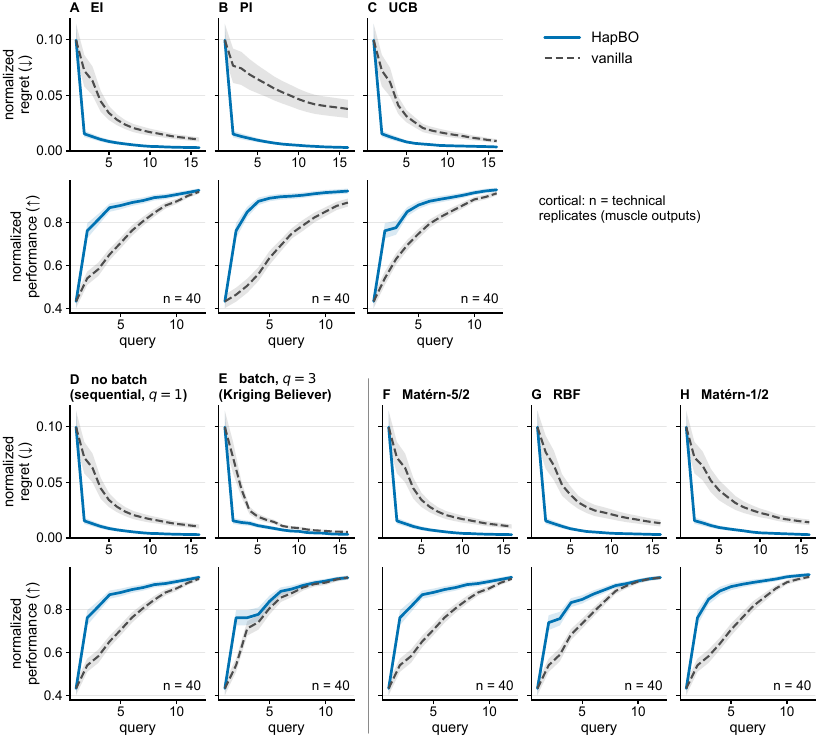}
\caption{\textbf{Bounded shaping across BO variants} (matched arms; only the stated ingredient
varies; details and conditions in the text of this appendix).  \textbf{A--C, acquisition
functions:} matched \hapbo{} vs.\ vanilla BO under EI, PI, and UCB; top rows HPOBench
(normalized simple regret, lower is better; 40 tasks), bottom rows cortical (normalized
performance, higher is better; $n=40$ technical replicates (muscle outputs); human+LLM prior, budget 12 queries); ten seeds.  \textbf{D--E, batch selection:}
the same matched EI contrast under sequential ($q{=}1$) and batch ($q{=}3$, Kriging Believer)
selection; D reproduces the EI panels of A.  \textbf{F--H, GP kernel:} EI-based
contrast with a Mat\'ern-5/2 (the kernel of A--E), RBF, or
Mat\'ern-1/2 kernel.}
\label{fig:variants}
\end{figure}

The bounded multiplier attaches to the acquisition landscape (it fundamentally re-sorts its ranking) and nothing else, so it should be
indifferent to the BO machinery around it: which acquisition is shaped, which kernel the GP surrogate uses, and whether queries are
issued one at a time or in batches.  This appendix verifies all three on matched pairs;
Figure~\ref{fig:variants} collects the evidence (panels A--C: acquisition functions; D--E: batch
selection; F--H: GP kernel).

\paragraph{Acquisition functions (Fig.~\ref{fig:variants}A--C).}
The bounded multiplier applies to any nonnegative base acquisition.  Panels A--C repeat the matched prior-on/prior-off contrast under EI, PI, and UCB on HPOBench and the cortical tasks.  Every arm
shares the GP surrogate, candidate set, and matched seeded initialization
($n_{\text{initial}}=1$); only the prior differs.  Raw UCB is an absolute score that
can fall below the incumbent, so its $\alpha$-$\pi$ arm shapes the clipped improvement $\max(\mathrm{UCB}-y^{+},0)$ over the incumbent $y^{+}$, keeping its behavior consistent with the nonnegative EI and PI improvements.  Because the
paired arms share the same seeded first query, their query-1 values coincide exactly (normalized
regret $0.099$ on HPOBench, normalized performance $0.433$ on cortical), so every subsequent
divergence is attributable to shaping alone.  Across all three acquisitions the shaped arm improves
low-budget sample efficiency: on HPOBench the final normalized regret has a $3$--$10\times$ reduction from vanilla to \hapbo{}; on
cortical the advantage is largest early, but seen throughout.

\paragraph{Batch selection (Fig.~\ref{fig:variants}D--E).}
The multiplier is likewise indifferent to whether queries are issued one at a time or in parallel
batches (an often likely setting when several electrodes or configurations are evaluated together).  We
keep the matched EI contrast of panels A--C and replace sequential selection with a
Kriging-Believer batch of size $q=3$: each round takes the shaped-EI argmax, fantasizes its outcome
as the posterior mean, refits, and repeats until the batch is full, then evaluates all three points.
The first member of a batch sees exactly the sequential arm's data, so divergence appears only
inside a batch, where the surrogate is frozen across picks. The shaped arm's advantage is within the early queries, in both settings.

\paragraph{GP kernel (Fig.~\ref{fig:variants}F--H).}
Every arm above uses a Mat\'ern-5/2 kernel.  Panels F--H change only the GP covariance: the Mat\'ern-5/2 default (F, which reproduces
the EI panels of A), the infinitely smooth squared-exponential kernel (RBF, G), and the rough
Mat\'ern-1/2 (exponential) kernel (H).    On HPOBench \hapbo{} has a $3.5\times$, $4.3\times$, and $4.7\times$ reduction of regret.  On cortical all \hapbo{} arms display an early lead. 

\section{Elicitation comparison: elicitor quality vs.\ downstream performance}
\label{app:ladder}

\begin{figure}[h]
\centering
\includegraphics[width=\linewidth]{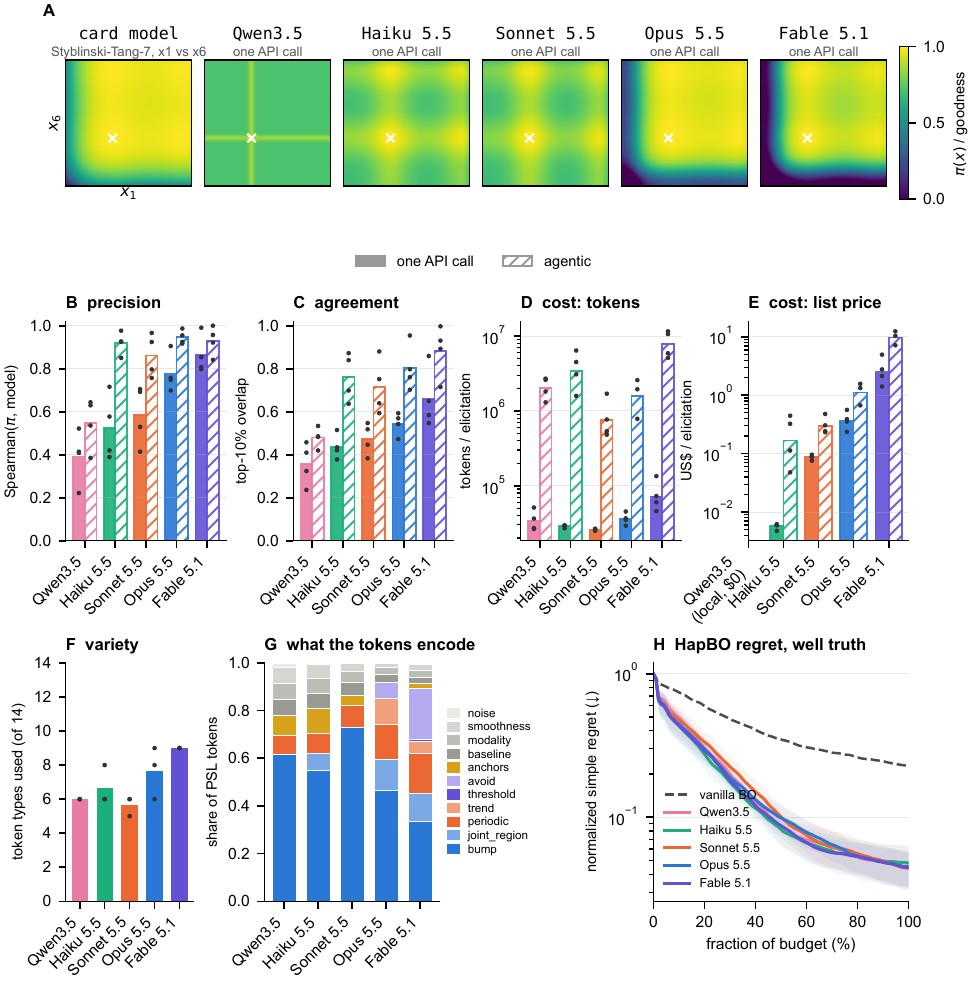}
\caption{\textbf{Downstream performance and elicitor quality.}  Synthetic panel, accurate cards
($\rho\approx0.9$); five elicitors and two elicitation routes (solid: one API call; hatched: agentic session).
\textbf{A}~Two-dimensional slice of the card model (Styblinski--Tang-7, $x_1$ vs.\ $x_6$, through its optimum, white
cross) and the compiled $\pi$ of each elicitor's single-call document on the same slice.  \textbf{B}~Precision
(rank correlation between $\pi$ and the card model) and \textbf{C}~agreement (overlap of the top 10\% of points); bars:
mean over the four functions, dots: functions.  \textbf{D}~Tokens and \textbf{E}~list-price cost per elicitation (the
local model has no marginal fee).  \textbf{F}~Number of distinct PSL token types used and \textbf{G}~share of each
token type in the single-call documents.  \textbf{H}~Normalized simple regret of \hapbo{} on the true function vs.\
fraction of budget, per elicitor (average of single-call and agentic curves, $\pm$ SEM); vanilla BO dashed.}  
\label{fig:ladder}
\end{figure}

Does a stronger elicitor produce a more useful prior?  We compare five elicitors of increasing
capability, the open-weights Qwen3.5-122B used throughout the paper and four commercial models
(Claude Haiku~5.5, Sonnet~5.5, Opus~5.5 and Fable~5.1), on the synthetic panel with the accurate cards
($\rho\approx0.9$).  Each elicitor transcribes the same card into PSL by two routes: a single API call and an agentic coding session in which the model can
compile and test its own document before submitting it (with a token cap of 2N, where N is the number of dimensions).  We
measure how faithfully the compiled prior $\pi$ reproduces the card model (precision and agreement), what the elicitation
costs, which PSL tokens the elicitor uses, and the downstream regret of \hapbo{} on the true function, with every
other setting held fixed (Fig.~\ref{fig:ladder}, Table~\ref{tab:ladder-cost}).

Stronger elicitors transcribe the card model more faithfully (Fig.~\ref{fig:ladder}A): the weaker
models draw a coarse, mostly separable structure, while the frontier models reproduce the correct landscape of the card
model, including precise depiction of favorable and unfavorable regions.  Precision and agreement increase with the capability of the model, and the agentic
route improves both at every level (B, C).  These gains come at a price that grows with both factors: the agentic
route requires far more tokens and a correspondingly higher cost than a single call, and more capable models cost more
per elicitation (D, E).  Moving up in model quality, elicitors tend to use a wider variety of PSL tokens (F) and change what
the tokens encode (G): weaker models rely mostly on single-dimension bumps, whereas stronger ones spread the document
over joint regions, periodic and trend components, and explicit \texttt{avoid} regions, stating also where the optimum
is not.  However, in this type of problem (H), elicitation with Qwen3.5 already captures the key features of the
search space robustly enough that all elicitors yield nearly overlapping regret curves, well below vanilla BO: benefits
in performance, however present, are limited.  In other optimization problems, for instance when the true optimum lies at the
second-, third- or lower-ranked candidate of the available model, further precision in the representation may have a
higher impact than in our synthetic problem suite.

\begin{table}[h]
\centering
\caption{\textbf{Elicitor comparison: downstream regret} Final normalized simple regret of
\hapbo{} on the true function at the end of the budget (lower is better), mean $\pm$ SEM of data in Fig.~\ref{fig:ladder}.   Bold marks the best value.}
\begin{tabular}{lc}
\toprule
Elicitor & final regret\\
\midrule
Qwen3.5 & 0.048 $\pm$ 0.015\\
Haiku 5.5 & 0.048 $\pm$ 0.016\\
Sonnet 5.5 & \textbf{0.044} $\pm$ 0.012\\
Opus 5.5 & 0.045 $\pm$ 0.012\\
Fable 5.1 & \textbf{0.044} $\pm$ 0.014\\
\midrule
vanilla BO & 0.228 $\pm$ 0.160\\
\bottomrule
\end{tabular}
\label{tab:ladder-cost}
\end{table}

\section{Targeted choices in baseline implementations}
\label{app:baseline-impl}

\begin{figure}[h]
\centering
\includegraphics[width=0.6\linewidth]{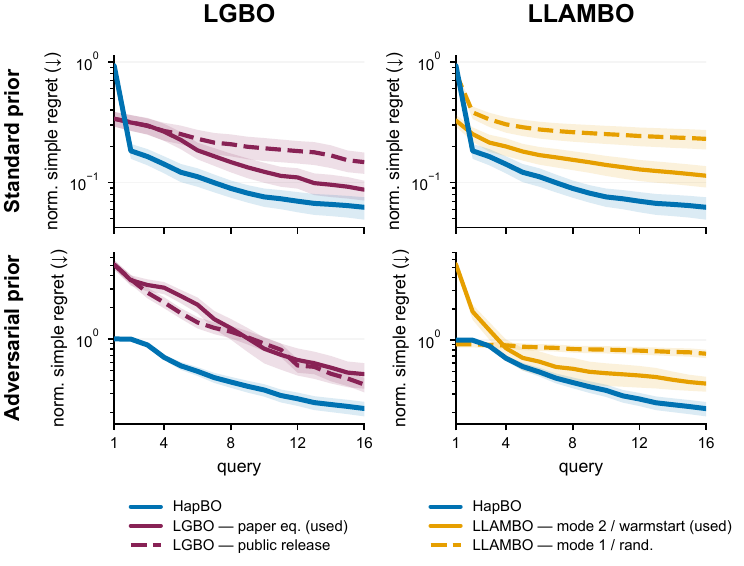}
\caption{\textbf{Baseline-implementation sensitivity.}  LGBO and LLAMBO, in two possible implementations, each shown with the matched \ours{} curve.  \textbf{Left, LGBO:} the
paper's mean-shift equation (solid, used in the main text) vs.\ the public release's confidence
operator (dashed).  \textbf{Right, LLAMBO:} random initialization, mode~1 (dashed) vs.\
zero-shot warm-start, mode~2 (solid, main text).  \textbf{Top:} standard prior, grid-span-normalized log regret.  \textbf{Bottom:} adversarial prior, vanilla-query-1-normalized regret (log). Bands are $\pm$SEM across tasks after averaging ten seeds.}
\label{fig:baseline-impl}
\end{figure}

Prior-guided and in-loop baselines admit more than one faithful implementation, and the choice can
move a curve as much as a change of method. Methods like LLAMBO and LGBO are designed to be modular and modified in future work. 
This is the case also for the current \hapbo{}, as discussed above, since the choice of $\alpha$ and metatokens for the surrogate is subject to welcome variations from future work.
For the two baselines LLAMBO and LGBO where we experienced the most open ambiguity at the time of integration, we
performed a direct comparison in a restricted HPOBench setting, holding the harness identical (same local model, task cards, GP where used, candidate set,
initialization, seeds, budget). Then, we adopted and reported in the main text the variant that performed best, so every
headline comparison is against the strongest competitor we could construct
(Fig.~\ref{fig:baseline-impl}).

\textbf{LLAMBO: initialization mode.}  The original paper presents random initialization (mode~1)
and zero-shot warm-starting with five LLM-proposed points (mode~2).  In our experiments, Mode~2
is markedly more sample-efficient through the low-budget regime; we therefore report mode~2.

\textbf{LGBO: confidence operator.}  No public code reproduces the LGBO (as defined in the original paper) on these benchmarks, so we
reconstruct it, and two readings exist: the paper's stated construction lifts the GP mean
by exactly $c$ posterior standard deviations (nonnegative, capped), whereas the public release maps
confidence through $\delta=3\,\Phi^{-1}(c)$, which is negative for $c<0.5$. This suppresses the very
region the model prefers, and is unbounded as $c\to1$.  We port both into the identical harness; the
paper-equation version is stronger on the standard prior and comparable on the adversarial prior; we therefore report our paper-equation version.

\section{Leak control}
\label{app:leak}

The objective of this paper is to determine, when LLM knowledge about a given problem exists, 
whether this can be integrated, by proper harnessing, in BO. 
However, there is a type of existing knowledge that is less desirable when testing any LLM-prior method on public benchmarks, which is contamination: a strong prior
might reflect recall of the final benchmark answer rather than transferable expertise.  
We address this with two
complementary controls: a direct probe of what the elicitor natively knows, and an additional objective whose
optimum cannot have been memorized.

\subsection{Does the elicitor natively know the optimum?}
\label{app:leak-probe}

For every task we present the task card and ask the model, from prior knowledge alone, for
its single best guess of the optimum's location.  We score the guess with the metric
\emph{recovered}\,\% $=(v-\bar v)/(v^\star-\bar v)$, the fraction of the head-room between the grid
mean and the optimum that the blind guess captures.  We read the reasoning trace and flag recall only when the model both names
the underlying benchmark and states its optimum.  For the analytic functions we probe the de-named
perturbed-model cards at all three fidelities, score against the optimum of the true (hidden-shift) objective, and add a
positive control that names the function outright.

No elicitor natively knows the optima from the cards alone
(Tables~\ref{tab:leak-probe}--\ref{tab:leak-probe-ladder}).  On HPOBench the blind guess recovers
$48$--$68\%$ of the head-room; on the cortical grid it
lands \emph{below} the map average ($-12\%$); no trace on any real task shows recall.  On the
analytic functions the guess tracks the fidelity of the supplied model and the de-named cards are
not recognized, with one exception: the elicitor recognizes the Ackley family at every fidelity yet
defers to the perturbed card rather than to a memorized optimum, so its guess still degrades with
$\rho$.  The named-function control recovers $100\%$ in every case, confirming the knowledge exists
but surfaces only when explicitly requested.  Across the four elicitors of Table~\ref{tab:leak-probe-ladder} the picture holds on all
five HPO families: a stronger model guesses marginally better blind (grand-mean recovered $60\%$
to $76\%$) but none of the 400 elicitations recalls an optimum.

\begin{table}[h]
\centering
\caption{\textbf{Probing native recall of benchmark optima.}  Recovered and raw \% of the blind best-guess optimum per
benchmark (local elicitor, greedy decoding), with reasoning-trace recall evidence.  HPO values are
means over the eight tasks per family ($\pm$SEM); cortical over the sixteen rat--muscle maps; SOLAR
is a single problem scored against the best feasible pool design.}
\resizebox{\linewidth}{!}{\begin{tabular}{lllll}
\toprule
Problem (family) & recovered \% & raw \% of best & recall evidence & partial recall evidence \\
\midrule
HPOBench lr & 60$\pm$11 & 98$\pm$1 & no recall evidence (0/8) & 0/8 \\
HPOBench rf & 48$\pm$9 & 94$\pm$1 & no recall evidence (0/8) & 0/8 \\
HPOBench xgb & 66$\pm$9 & 99$\pm$0 & no recall evidence (0/8) & 0/8 \\
HPOBench nn & 68$\pm$9 & 96$\pm$2 & no recall evidence (0/8) & 0/8 \\
HPOBench svm & 55$\pm$16 & 89$\pm$5 & no recall evidence (0/8) & 0/8 \\
NeuroMOSAICS cortical & -12$\pm$4 & 37$\pm$6 & no recall evidence (0/16) & 0/16 \\
SOLAR & 98 & 98 & no recall evidence (0/1) & 0/1 \\
\bottomrule
\end{tabular}} %
\label{tab:leak-probe}
\end{table}

\begin{table}[h]
\centering
\caption{\textbf{Probing native recall of benchmark optima, analytic functions.}  De-named cards supply a perturbed
model at three fidelities $\rho$; the guess is scored on the true (hidden-shift) objective
(recovered \% and normalized RMS distance to the true and to the textbook $x^\star$; $0=$ exact).  The named control names the
function outright.}
\resizebox{\textwidth}{!}{\begin{tabular}{llrrrp{5.2cm}}
\toprule
Function & card & recovered \% & dist.\ to true $x^\star$ & dist.\ to textbook $x^\star$ & recall evidence \\
\midrule
Hartmann-6 & $\rho{\approx}0.9$ & 86 & 0.47 & 0.49 & follows provided model, no family recognition \\
 & $\rho{\approx}0.5$ & 7 & 0.23 & 0.27 & follows provided model, no family recognition \\
 & $\rho{\approx}0.1$ & -3 & 0.31 & 0.37 & follows provided model, no family recognition \\
 & named control & 100 & 0.00 & 0.00 & names Hartmann, states $x^\star$ \\
\midrule
Styblinski--Tang-7 & $\rho{\approx}0.9$ & 98 & 0.03 & 0.11 & follows provided model, no family recognition \\
 & $\rho{\approx}0.5$ & 41 & 0.44 & 0.50 & follows provided model, no family recognition \\
 & $\rho{\approx}0.1$ & -370 & 0.53 & 0.54 & follows provided model, no family recognition \\
 & named control & 100 & 0.00 & 0.00 & names Styblinski--Tang, states $x^\star$ \\
\midrule
Levy-10 & $\rho{\approx}0.9$ & 99 & 0.03 & 0.20 & follows provided model, no family recognition \\
 & $\rho{\approx}0.5$ & 73 & 0.17 & 0.30 & follows provided model, no family recognition \\
 & $\rho{\approx}0.1$ & -118 & 0.41 & 0.45 & follows provided model, no family recognition \\
 & named control & 100 & 0.00 & 0.00 & names Levy, states $x^\star$ \\
\midrule
Ackley-15 & $\rho{\approx}0.9$ & 97 & 0.00 & 0.23 & recognizes Ackley family, guess follows the provided (shifted) model \\
 & $\rho{\approx}0.5$ & 6 & 0.20 & 0.31 & recognizes Ackley family, guess follows the provided (shifted) model \\
 & $\rho{\approx}0.1$ & -1 & 0.51 & 0.50 & recognizes Ackley family, guess follows the provided (shifted) model \\
 & named control & 100 & 0.00 & 0.00 & names Ackley, states $x^\star$ \\
\bottomrule
\end{tabular}}
\label{tab:leak-probe-synth}
\end{table}

\begin{table}[h]
\centering
\caption{\textbf{Probing native recall of benchmark optima across models.}  Recovered \% of the blind
best-guess per HPOBench family (Claude elicitors at temperature $1.0$, three draws per task; local
elicitor greedy).  No elicitation on any family recalls an optimum.}
\resizebox{\textwidth}{!}{\begin{tabular}{lccccccll}
\toprule
Elicitor & lr & rf & xgb & nn & svm & grand mean & recall evidence & partial recall evidence \\
\midrule
Qwen3.5-122B & 60$\pm$11 & 48$\pm$9 & 66$\pm$9 & 68$\pm$9 & 55$\pm$16 & 60$\pm$5 & no recall evidence (0/40) & 0/40 \\
Claude Opus & 75$\pm$3 & 75$\pm$4 & 74$\pm$3 & 80$\pm$4 & 76$\pm$4 & 76$\pm$2 & no recall evidence (0/120) & 2/120$^{\dagger}$ \\
Claude Sonnet & 69$\pm$5 & 67$\pm$5 & 72$\pm$4 & 80$\pm$3 & 80$\pm$5 & 74$\pm$2 & no recall evidence (0/120) & 1/120$^{\dagger}$ \\
Claude Haiku & 63$\pm$6 & 50$\pm$5 & 74$\pm$3 & 74$\pm$6 & 67$\pm$6 & 65$\pm$2 & no recall evidence (0/120) & 0/120 \\
\bottomrule
\end{tabular}}
\label{tab:leak-probe-ladder}
\end{table}

\subsection{An un-memorizable CNN benchmark}
\label{app:leak-cnn}

\begin{figure}[h]
\centering
\begin{minipage}[b]{0.54\linewidth}\centering
\includegraphics[width=\linewidth]{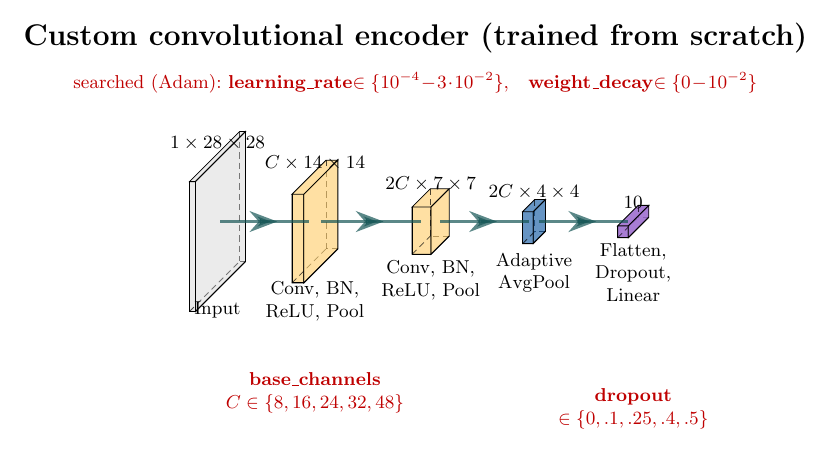}\\[3pt]{\small (A)}
\end{minipage}\hfill
\begin{minipage}[b]{0.44\linewidth}\centering
\includegraphics[width=\linewidth]{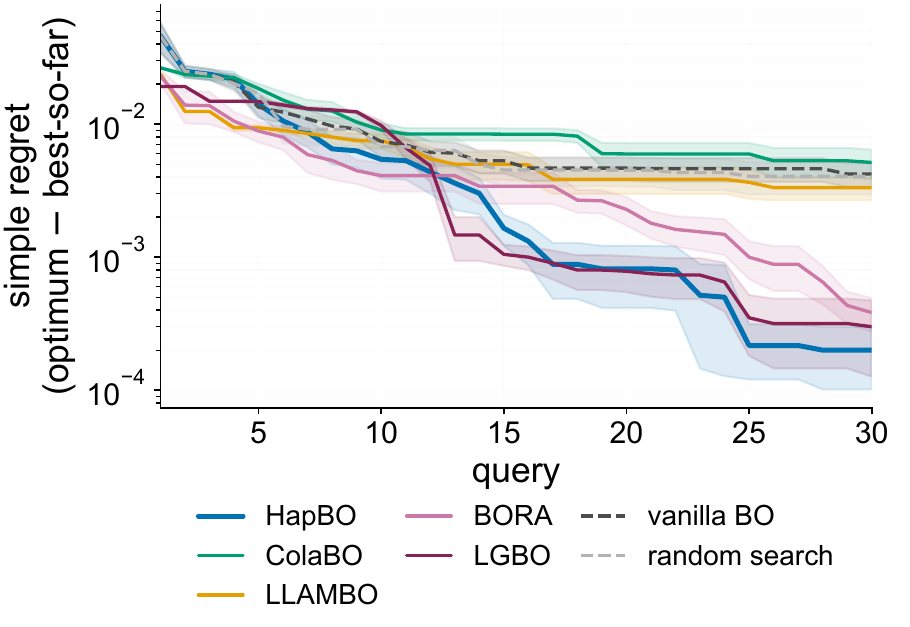}\\[3pt]{\small (B)}
\end{minipage}
\caption{\textbf{Leak-control benchmark.}  \textbf{(A)} the bespoke convolutional encoder optimized
as an un-memorizable objective, with the four searched hyperparameters highlighted.
\textbf{(B)} query-wise simple regret (log axis, mean $\pm$ SEM over ten
seeds) on the 750-configuration grid.  The optimum is a property of our network, grid, and training
protocol and appears in no public corpus, so no method's prior can reflect memorization.}
\label{fig:leak-control}
\end{figure}

We define a small bespoke convolutional encoder (Fig.~\ref{fig:leak-control}A) and compute its
hyperparameter-response surface ourselves: a 4-D grid of $6\times5\times5\times5=750$
configurations over learning rate, base channel width, dropout, and weight decay, each network
trained from scratch and scored by seed-averaged validation accuracy on a fixed small image
subsample.
This exact architecture-and-grid response map exists in no public corpus, so the optimum is not
recoverable from memory; a competent elicitor can still place a reasonable prior from general
machine-learning intuition, which is the ability we mean to isolate.  On this objective \ours{} reaches the lowest simple regret, ahead of LGBO, BORA, whereas prior-free controls, LLAMBO and ColaBO clearly trail.

\section{Further baselines}
\label{app:cake}

\begin{figure}[h]
\centering
\includegraphics[width=\linewidth]{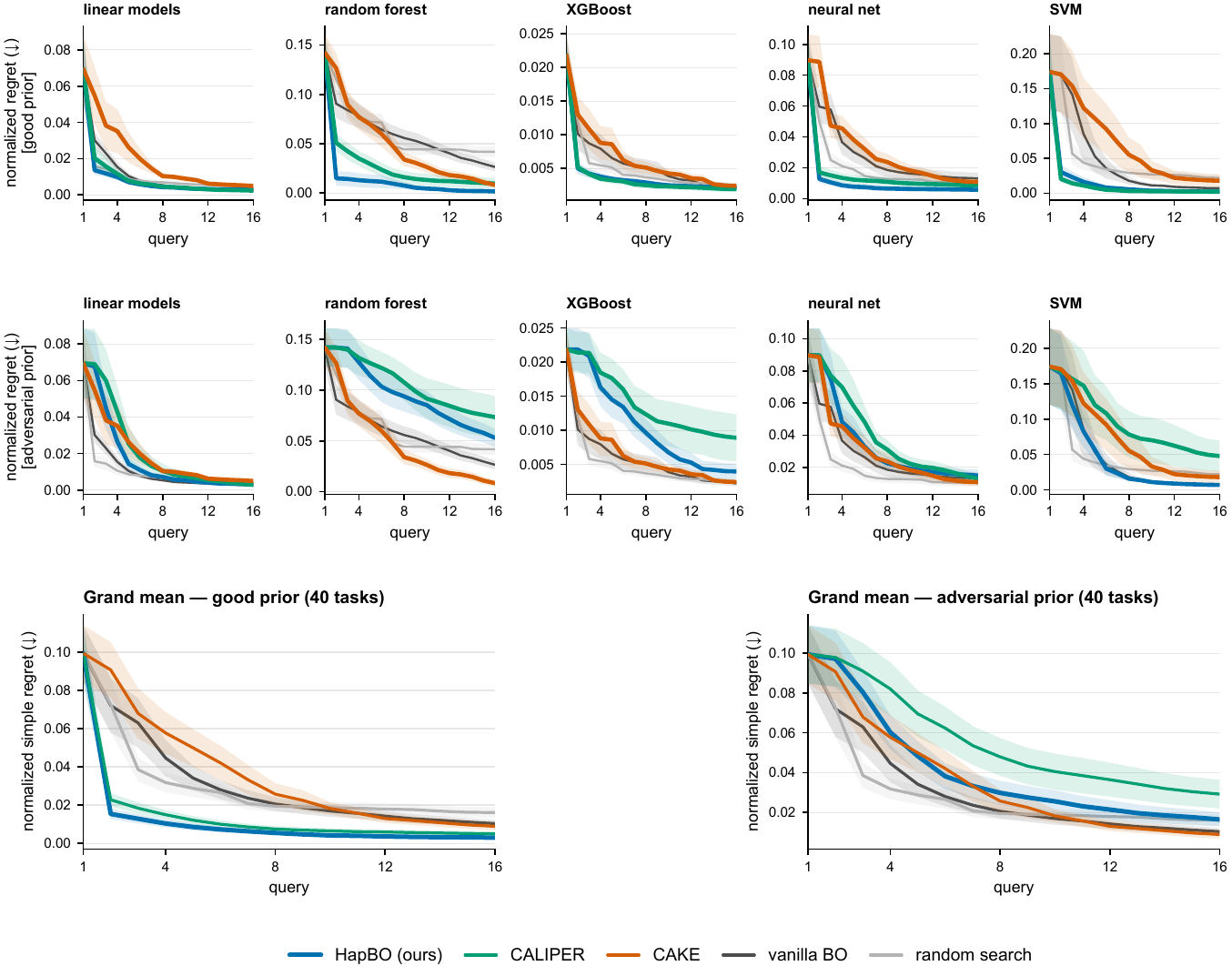}
\caption{\textbf{CAKE and CALIPER vs.\ \hapbo{} on HPOBench.}  Normalized simple regret
(lower is better) per model family under a faithful prior (top) and an adversarial prior
(middle), with grand means over the 40 tasks (bottom; faithful left, adversarial right).  All arms
share the GP surrogate and matched random initialization across ten seeds; CAKE is
prior-agnostic, so the same curve appears in both conditions.}
\label{fig:cake-caliper}
\end{figure}

Two recent LLM-driven methods keep the model in the loop in ways orthogonal to the prior-weighting
baselines of the main text; we run both in our harness on the HPOBench suite
(five families, ten seeds), under the faithful and the  adversarial prior.
\textbf{CAKE} \citep{NEURIPS2025_c03a2610} uses the LLM as a crossover/mutation operator that
evolves the GP kernel expression at each query; it consumes no compiled belief, so its curve is
identical in both prior conditions.  \textbf{CALIPER}
\citep{chen2026evidencegatedllmpriorsmultiobjective} elicits per-role priors once and admits them
through a discrete counterfactual evidence gate; we adapt it to single-objective HPO (three
objective personas; EI in place of the hypervolume acquisition) and, absent public code,
reimplement it from the paper.  \ours{} attains the lowest grand-mean regret on the good prior, slightly ahead of CALIPER, which however drops largely with an adversarial prior
(Fig.~\ref{fig:cake-caliper}) whereas the bounded multiplier keeps \ours{} robust; CAKE always performs in line with vanilla BO.

\begin{figure}[h]
\centering
\includegraphics[width=\linewidth]{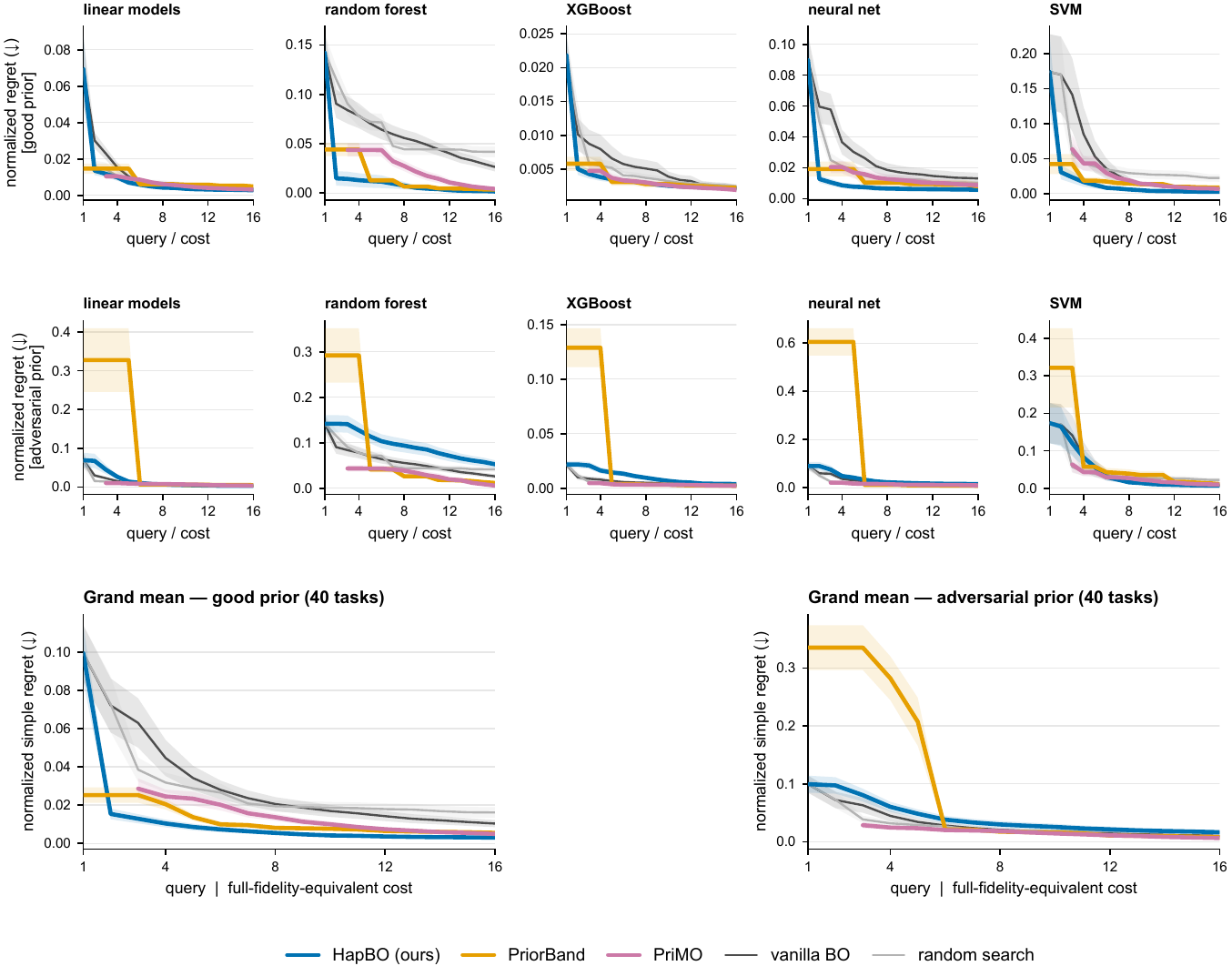}
\caption{\textbf{PriorBand and PriMO vs.\ \hapbo{} on HPOBench.}  Normalized simple regret
(lower is better) per model family under a faithful prior (top) and an adversarial prior (middle), with grand
means over the 40 tasks (bottom; faithful left, adversarial right); ten seeds.  PriorBand and PriMO are plotted
against full-fidelity-equivalent cost (budget 16), the other arms against queries.}
\label{fig:priorband-primo}
\end{figure}

Two expert-prior HPO methods take the belief as one value and one confidence level per hyperparameter, which
we elicit with the same LLM from the same cards as \ours{}.  \textbf{PriorBand} \citep{NEURIPS2023_1704fe7a}
and \textbf{PriMO} \citep{basu2025primo} (official NePS~0.17.0 implementations, used with original
papers' settings) are multi-fidelity and the only methods in this paper with access to cheap low-fidelity
evaluations: they also evaluate cheaper versions of each task (fewer iterations or trees, a data subsample), each charged $z/z_{\max}$ of a full evaluation within the same budget of 16, and are
scored by their best full-fidelity evaluation.  Both are less performant than \ours{} under the same good
LLM-elicited prior and substantially on par under an adversarial prior (Fig.~\ref{fig:priorband-primo}).

\begin{figure}[h]
\centering
\includegraphics[width=\linewidth]{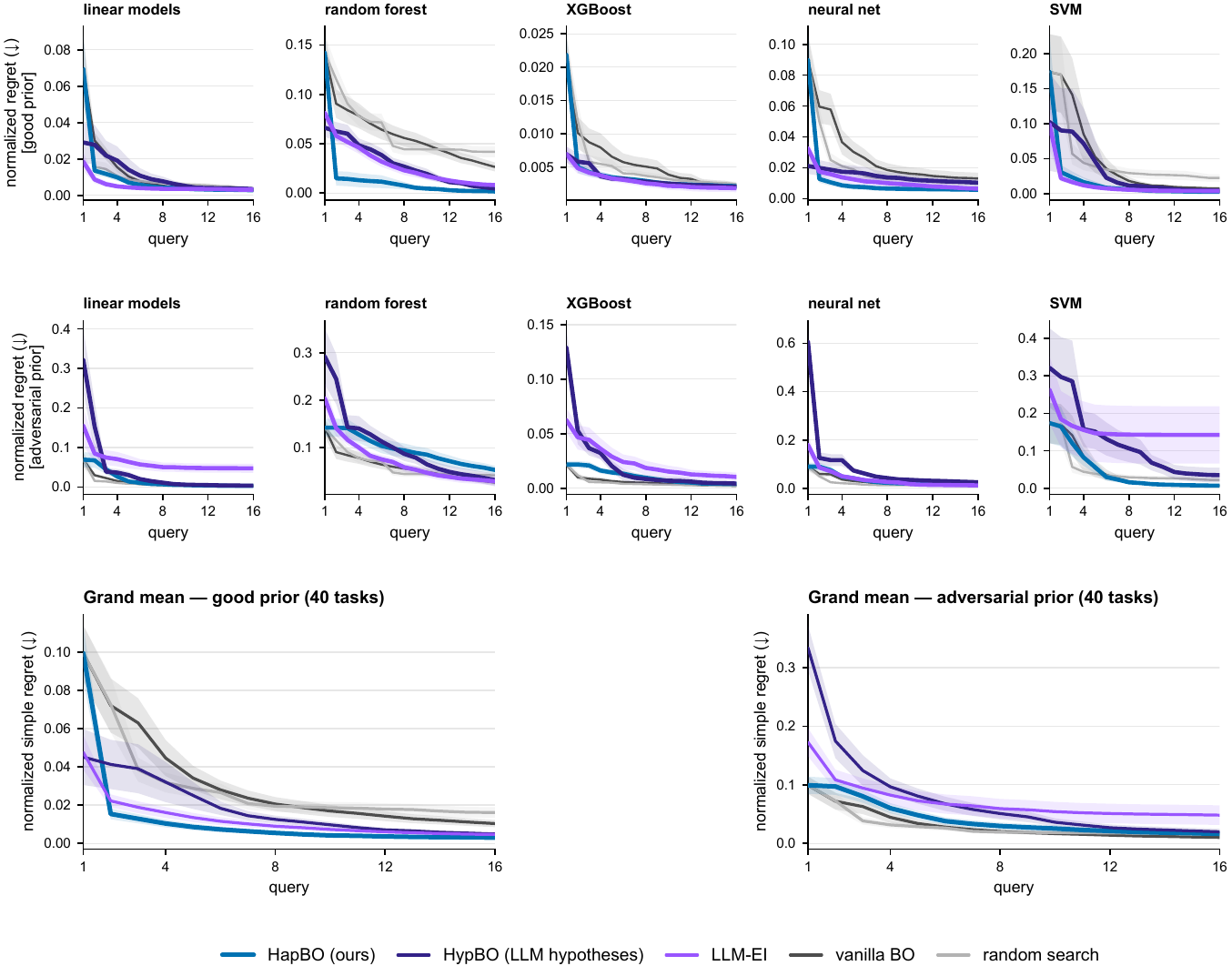}
\caption{\textbf{HypBO and LLM-EI vs.\ \hapbo{} on HPOBench.}  Normalized simple regret
(lower is better) per model family under a faithful prior (top) and an adversarial prior (middle), with grand
means over the 40 tasks (bottom; faithful left, adversarial right).  All arms run for ten
seeds; HypBO and LLM-EI read the same task card as \hapbo{}.}
\label{fig:hypbo-llmei}
\end{figure}

Two further methods use the LLM's knowledge as something other than a weighted prior.  \textbf{HypBO}
\citep{ijcai2024p429} searches boxes around optimum guesses that the LLM states with a confidence, elicited as in
BORA \citep{ijcai2025p553}; we run its public code unmodified, with one initial point (its authors use five).
\textbf{LLM-EI} \citep{patel2026llmei} fits a preference GP to the LLM's answers to a pairwise survey of each card
(4--8k questions per task) and restricts EI to the top 15\% of the resulting utility.  Both trail \ours{} under both faithful and adversarial priors
(Fig.~\ref{fig:hypbo-llmei}).

\section{Theory}
\label{app:theory}

This appendix contains the formal analysis behind \hapbo{}: the expressivity and organization of
the Prior Specification Language (PSL), the acquisition-level guarantee of the bounded multiplier,
and two separate convergence analyses: 1) noiseless fixed-kernel EI and 2) noisy finite-domain GP-UCB.
The two convergence regimes are analyzed independently.

\textbf{Notation.}  The search space $\mathcal X=\mathcal X_1\times\cdots\times\mathcal X_d$ is a
typed product: each coordinate is a compact interval after the schema's scale transform (e.g.\
log), a finite ordinal set, or a finite categorical set; $\mathcal X$ is compact, and a
finite-candidate implementation acts on a grid $D=\{x_1,\dots,x_N\}\subseteq\mathcal X$.  The
objective is $f$, maximized unless stated otherwise; $X^\star=\arg\max_x f(x)$ (nonempty on
compact $\mathcal X$ for continuous $f$, always on finite $D$).

\textbf{Three regret quantities.}  We distinguish:
\begin{itemize}
\item the \emph{instantaneous regret} $r_t=f(x^\star)-f(x_t)$ of the point queried at round $t$;
\item the \emph{best-sampled latent simple regret} $S_T=\min_{t\le T} r_t$, the regret of the best
point \emph{queried} within the budget, evaluated on the latent objective (under noise, $f(x_t)$ is
not observed exactly, so $S_T$ is not directly observable);
\item the \emph{recommendation regret} $f(x^\star)-f(\hat x_T)$ of a point $\hat x_T$ recommended
from a noisy history.
\end{itemize}
Each result below controls exactly one of these: the EI theorem controls best-sampled simple
regret in a noiseless setting (where it is observable); the UCB theorem controls cumulative
instantaneous regret $R_T=\sum_{t\le T}r_t$, and through $S_T\le R_T/T$ the best-sampled latent
simple regret.  

\subsection{PSL compiler: expressivity and vocabulary structure}
\label{app:theory-psl}

The results of this subsection characterize the compiler independently of any optimizer.

\subsubsection{Typed domain and compiler contract}
\label{app:theory-contract}

A PSL document $P$ (typed schema; primitives: anchors, bumps, joint regions, trends, thresholds,
periodic, ridges, conditionals, avoid regions, plateaus, categorical weights; trust metadata)
compiles as $\mathcal C(P)=(\pi_P,\alpha_P,F_P)$, where $\pi_P:\mathcal X\to[0,1]$ is the prior
belief function, $\alpha_P\le\alpha_{\max}<1$ is the compiled trust, and $F_P\subseteq\mathcal X$
is the feasible set.  The semantics clauses used by the theory:

\begin{itemize}
\item[\textbf{(S1)}] \emph{Finite anchors.}  On finite $D$, an anchor at $x_i$ with value
$p_i\in[0,1]$ sets the compiled prior at $x_i$ to exactly $p_i$, with no renormalization or
clipping; candidates not covered by any anchor take the baseline value.
\item[\textbf{(S2)}] \emph{Continuous anchors.}  On continuous coordinates, for any finite anchor
set $z_1,\dots,z_K$ forming a $\delta$-cover, the compiler realizes a continuous partition of unity
subordinate to that cover: $\pi_P(x)=\sum_j v_j\psi_j(x)$ with $\psi_j\ge0$, $\sum_j\psi_j=1$, and
$\psi_j(x)>0$ only when $\operatorname{dist}(x,z_j)<\delta$.
\item[\textbf{(S3--S6)}] \emph{Macro semantics.}  The baseline compiles to an affine function; a
trend to a piecewise-linear saturating ramp; a periodic token to a sinusoidal envelope mapped into
$[0,1]$; a bump or avoid to a signed Gaussian $\pm w\exp(-(x-\mu)^2/(2\sigma^2))$ with weight
$w\in[0,1]$; a joint region to the corresponding multi-dimensional product.
\item[\textbf{(S7)}] \emph{Feasibility.}  No belief primitive removes a candidate.  Hard limits
are part of \emph{defining} the search space: they enter through the schema's typed per-dimension
ranges (and, for a residual joint limit, a caller-supplied feasibility filter), fixing $F_P$ before
any belief is compiled.  The soft \texttt{avoid} primitive subtracts from the score, which is then
clamped to $[0,1]$; it can lower a region's plausibility but never removes it from $F_P$.
\end{itemize}

\begin{theorem}[PSL expressivity]
\label{thm:psl-expressivity}
Let the compiler satisfy (S1)--(S2).
\begin{itemize}
\item[(a)] \textbf{Finite exactness.}  For every $p\in[0,1]^N$ on a finite candidate set
$D=\{x_1,\dots,x_N\}$ there is an anchors-only PSL document $P$ with $\pi_P(x_i)=p_i$ for all $i$.
\item[(b)] \textbf{Continuous density.}  For every continuous $g:\mathcal X\to[0,1]$ (categorical
dimensions as finite strata) and every $\varepsilon>0$ there is an anchors-only document $P$ with
$\sup_{x\in\mathcal X}|\pi_P(x)-g(x)|<\varepsilon$.
\item[(c)] \textbf{Quantitative rate.}  If the continuous block of a stratum satisfies
$S\subseteq\prod_{i=1}^{d_c}[0,D_i]$ and $g$ is $L$-Lipschitz there, accuracy $\varepsilon$ is
achieved with
$K\le\prod_i\lceil D_i\sqrt{d_c}\,L/\varepsilon\rceil$
anchors per stratum; equivalently, the sup-error in the anchor budget $K$ is $O(K^{-1/d_c})$.
\end{itemize}
\end{theorem}

\begin{proof}
(a)~One anchor per candidate with value $p_i$; clause (S1) returns the table exactly, and each
$p_i\in[0,1]$ needs no clipping.

(b)~Fix a continuous stratum $S$ (compact).  Uniform continuity gives $\delta>0$ with
$\operatorname{dist}(x,y)<\delta\Rightarrow|g(x)-g(y)|<\varepsilon$.  Take a finite $\delta$-cover
$z_1,\dots,z_K$ of $S$ and the partition of unity of clause (S2); set anchor values $g(z_j)$.  Then
\begin{equation*}
|\pi_P(x)-g(x)|=\Bigl|\sum_j\psi_j(x)\bigl(g(z_j)-g(x)\bigr)\Bigr|
\le\sum_j\psi_j(x)\,|g(z_j)-g(x)|<\varepsilon,
\end{equation*}
since $\psi_j(x)>0$ forces $\operatorname{dist}(x,z_j)<\delta$ and the $\psi_j$ sum to one.
Repeat per categorical stratum, gating anchor sets by the categorical assignment.  The interpolant
is a convex combination of values in $[0,1]$, hence a valid prior field.

(c)~Partition the bounding box into axis-aligned cells of side $\delta/\sqrt{d_c}$ with
$\delta=\varepsilon/L$; each cell has diameter $\delta$, so anchors placed one per cell meeting $S$
form a $\delta$-cover of size at most the stated product.  Running the argument of (b) with this
cover bounds the sup-error by $L\delta=\varepsilon$; inverting the cell count gives the
$O(K^{-1/d_c})$ form.
\end{proof}

\begin{proposition}[vocabulary organization]
\label{prop:psl-anova}
Let $(\mathcal X,\mu)$ be a product probability space and $g\in L^2(\mathcal X,\mu)$.  Define
recursively $g_\emptyset=\mathbb E[g]$ and
$g_S(x_S)=\mathbb E[g\mid x_S]-\sum_{T\subsetneq S}g_T(x_T)$.  Then
$g=\sum_{S\subseteq\{1,\dots,d\}}g_S$ in $L^2$ with $\langle g_S,g_T\rangle=0$ for $S\ne T$, and
the order-2 truncation $g_{\le2}=g_\emptyset+\sum_i g_i+\sum_{i<j}g_{ij}$ satisfies
\begin{equation*}
\|g-g_{\le2}\|_2^2=\sum_{|S|>2}\|g_S\|_2^2 .
\end{equation*}
PSL's layers mirror this decomposition: the baseline is the order-0 component, the marginal
primitives the order-1 components, the interaction primitives the order-2 components, and anchors
the residual complete basis.
\end{proposition}

\begin{proof}
The recursion gives $\mathbb E[g_S\mid x_T]=0$ for every strict subset $T\subsetneq S$; the product
structure of $\mu$ then yields orthogonality across distinct index sets, and the recursive sum
telescopes to $g$.  For the truncation, $g-g_{\le2}=\sum_{|S|>2}g_S$ and orthogonality gives the
Pythagorean identity.
\end{proof}

\emph{Limitation.}  Proposition~\ref{prop:psl-anova} explains the vocabulary's organization; it
does not show that low-order structure dominates any particular elicited prior.

\begin{remark}[macros cannot reduce expressivity]
Enlarging the vocabulary with macro tokens cannot reduce what is expressible: any document written
in a sub-vocabulary is a document of the full vocabulary, so exactness and density
(Theorem~\ref{thm:psl-expressivity}) are retained.  The macro tokens exist for economy of
expression, not for additional expressive power.
\end{remark}

\subsection{Common bounded-shaping results}
\label{app:theory-shaping}

\begin{definition}[raw bounded shaping]
\label{def:shaping}
Let $A_t:F_P\to\mathbb R$ be a base acquisition at round $t$.  The shaped acquisition is
\begin{equation}
A_t^M(x)=A_t(x)\,M_t(x),\qquad
M_t(x)=1-\alpha_t+\alpha_t\,\pi_t(x),\qquad
m:=1-\alpha_{\max}>0,
\label{eq:shaping}
\end{equation}
with $\pi_t(x)\in[0,1]$ and $\alpha_t\in[0,\alpha_{\max}]$.  The default is a run-fixed field
$\pi$ and scalar static trust $\alpha$ compiled before the run; time-varying trust
$\alpha_t$ is introduced only for the decay results below.  The implementation clamps
$\alpha_{\max}=0.9$, so $1/m\le10$.  We apply the multiplier to the \emph{raw} acquisition, unlike Eq.~10 of \citealp{GUAYHOTTIN2025113039}, which min--max normalizes the acquisition before multiplying. These generally define different selection rules: raw shaping preserves positive acquisition support, while min--max normalization can assign zero to candidates with positive raw acquisition. Nevertheless, the bounded-influence and EI regret guarantees also hold for the normalized formulation (Corollary~\ref{cor:normalized-shaping}).
\end{definition}

\begin{proposition}[bounded influence on acquisition selection]
\label{prop:shaping}
For $M_t$ as in Eq.~\ref{eq:shaping}: $m\le M_t(x)\le1$ for all $x$ and $t$.  Consequently, if
$A_t\ge0$ on the feasible set and $x_t\in\arg\max_x A_t(x)M_t(x)$, then
\begin{equation*}
A_t(x_t)\ \ge\ m\,\max_x A_t(x).
\end{equation*}
\end{proposition}

\begin{proof}
$M_t=1-\alpha_t(1-\pi_t)$ with $\alpha_t(1-\pi_t)\in[0,\alpha_{\max}]$ gives the bounds.  For the
selection statement, $A_t(x_t)\ge A_t(x_t)M_t(x_t)\ge A_t(x)M_t(x)\ge m\,A_t(x)$ for every $x$,
using $M_t(x_t)\le1$, optimality of $x_t$, and $M_t(x)\ge m$.
\end{proof}

\subsection{Regime I: noiseless fixed-kernel expected improvement}
\label{app:theory-ei}

This subsection follows the minimization convention of the classical EI analysis
\citep{JMLR:v12:bull11a}: $f$ is minimized (equivalently $-f$ is maximized); the statements are orientation-free.  $L_n$ denotes
the nonnegative best-sampled simple regret after $n$ evaluations, observable here because
observations are noiseless.

\begin{assumption}[EI regime]
\label{assump:ei}
(i)~Observations are deterministic and noiseless; (ii)~the domain is compact; (iii)~the kernel,
lengthscale, and scale are fixed during the run, with smoothness $\nu<\infty$; (iv)~the objective
lies in the RKHS ball $\|f\|_{\mathcal H_\ell(\mathcal X)}\le R$; (v)~the acquisition is raw EI,
shaped as in Definition~\ref{def:shaping} with a run-fixed multiplier $M\in[m,1]$, continuous in
$x$ so the shaped argmax exists; (vi)~no sequential GP hyperparameter refitting.  In addition, the
two selection-dependent ingredients of the classical machinery hold for \emph{any} selection
sequence: the covering argument of \citet[Lemma~7]{JMLR:v12:bull11a} bounds the posterior
standard deviation at all but at most $p$ rounds by $C\,p^{-(\nu\wedge1)/d}\log^\kappa p$, and the
running-optimum telescoping bounds the incumbent decrement $z^*_n-f(x_{n+1})>2Rp^{-1}$ at no more
than $p$ rounds; a pigeonhole over the window $[p,3p]$ yields an index $n_p\in[p,3p]$ satisfying
both bounds simultaneously.  Finally, EI satisfies the two-sided sandwich
\citep[Lemma~8]{JMLR:v12:bull11a}: with $I=(f(x^*_n)-f(x))^+$, $s=s_n(x;\ell)$, and
$\tau(z)=z\Phi(z)+\phi(z)$,
\begin{equation}
\max\Bigl(I-Rs,\ \tfrac{\tau(-R/\rho)}{\tau(R/\rho)}\,I\Bigr)\ \le\ \mathrm{EI}_n(x)\ \le\
I+(R+\rho)\,s .
\label{eq:ei-sandwich}
\end{equation}
\end{assumption}

For a run-fixed multiplier $M$ define
\begin{equation}
M(\mathrm{opt}):=\max_{x^\star\in X^\star}M(x^\star),\qquad
\Lambda(M):=\frac{\sup_x M(x)}{M(\mathrm{opt})}\ \le\ \frac{1}{m},
\label{eq:lambda}
\end{equation}
using $\sup_xM\le1$ and $M(\mathrm{opt})\ge m$; note $\Lambda(M)\ge1$ always.

\begin{theorem}[shaped-EI rate preservation]
\label{thm:ei}
Under Assumption~\ref{assump:ei}, the best-sampled simple regret of shaped EI satisfies, for
$3p\le n<3(p+1)$ with $p$ past the initial design,
\begin{equation}
L_n(\mathrm{EI}^M)\ \le\ \Lambda(M)\,
\frac{\tau(R/\rho)}{\tau(-R/\rho)}
\Bigl(2Rp^{-1}+C(R+\rho)\,p^{-(\nu\wedge1)/d}\log^\kappa p\Bigr)
\ =\ \Lambda(M)\,B_n(\mathrm{EI}),
\label{eq:ei-rate}
\end{equation}
where $B_n(\mathrm{EI})=O\bigl(n^{-(\nu\wedge1)/d}\log^\kappa n\bigr)$ denotes the Bull-style
theoretical upper bound on the loss of unshaped EI---an upper bound, not the realized loss of a
separately executed vanilla run.  In particular $L_n(\mathrm{EI}^M)\le(1/m)\,B_n(\mathrm{EI})$:
bounded shaping preserves the classical rate, with a constant determined by the multiplier at an
optimizer.
\end{theorem}

\begin{proof}
Fix $p$ past the initial design and let $n_p\in[p,3p]$ be the pigeonhole index of
Assumption~\ref{assump:ei}.  Let $x^\star$ be a global minimizer maximizing $M$ over $X^\star$.
By the lower branch of \eqref{eq:ei-sandwich} at $x^\star$, with
$I_{n_p}(x^\star)=z^*_{n_p}-z^*$,
\begin{equation*}
\mathrm{EI}_{n_p}(x^\star)\ \ge\ \frac{\tau(-R/\rho)}{\tau(R/\rho)}\,(z^*_{n_p}-z^*).
\end{equation*}
By optimality of the shaped selection at step $n_p$,
$\mathrm{EI}_{n_p}(x^\star)M(x^\star)\le\mathrm{EI}_{n_p}(x_{n_p+1})M(x_{n_p+1})$, hence
\begin{equation*}
\mathrm{EI}_{n_p}(x^\star)\ \le\ \mathrm{EI}_{n_p}(x_{n_p+1})\,
\frac{M(x_{n_p+1})}{M(x^\star)}\ \le\ \Lambda(M)\,\mathrm{EI}_{n_p}(x_{n_p+1}).
\end{equation*}
By the upper branch of \eqref{eq:ei-sandwich} at $x_{n_p+1}$ and the two pigeonhole bounds,
\begin{equation*}
\mathrm{EI}_{n_p}(x_{n_p+1})\ \le\ \bigl(z^*_{n_p}-f(x_{n_p+1})\bigr)^++(R+\rho)\,s_{n_p}(x_{n_p+1})
\ \le\ 2Rp^{-1}+C(R+\rho)\,p^{-(\nu\wedge1)/d}\log^\kappa p .
\end{equation*}
Chaining the three displays and using monotonicity $z^*_n\le z^*_{n_p}$ for $n\ge n_p$ (in
particular for $3p\le n<3(p+1)$, since $n_p\le3p$) gives \eqref{eq:ei-rate}.
\end{proof}

\begin{corollary}[Min--max-normalized acquisition shaping]
\label{cor:normalized-shaping}
Let $A_t\ge0$, $M_t\in[m,1]$, and define
$a_t=\min_x A_t(x)$ and $b_t=\max_x A_t(x)$.
For $b_t>a_t$, consider the normalized selection rule
\begin{equation*}
x_t\in\arg\max_x
\frac{A_t(x)-a_t}{b_t-a_t}\,M_t(x).
\end{equation*}
Then $A_t(x_t)\ge m\,b_t$, as in
Proposition~\ref{prop:shaping}. Moreover, for EI under
Assumption~\ref{assump:ei}, with the raw selection
rule in (v) replaced by this normalized rule,
the conclusion of Theorem~\ref{thm:ei} holds
with the same constant $\Lambda(M)$.
\end{corollary}

\begin{proof}
By optimality of the normalized selection, for every $x$,
\begin{equation*}
(A_t(x_t)-a_t)M_t(x_t)
\ge (A_t(x)-a_t)M_t(x).
\end{equation*}
Taking $x\in\arg\max A_t$ and using
$m\le M_t\le1$ gives
\begin{equation*}
A_t(x_t)\ge a_t+m(b_t-a_t)
\ge m\,b_t,
\end{equation*}
since $a_t\ge0$.

For EI, let $x^\star$ be a global optimizer
maximizing $M$ over $X^\star$. The same
optimality inequality yields
\begin{equation*}
A_t(x^\star)-a_t
\le \frac{M(x_t)}{M(x^\star)}
(A_t(x_t)-a_t)
\le \Lambda(M)(A_t(x_t)-a_t).
\end{equation*}
Since $a_t\ge0$ and $\Lambda(M)\ge1$,
\begin{equation*}
A_t(x^\star)\le\Lambda(M)A_t(x_t).
\end{equation*}
This is exactly the selection inequality
used in the proof of Theorem~\ref{thm:ei};
all subsequent steps are unchanged.
When $b_t=a_t$, both acquisition comparisons
hold trivially, irrespective of the selected point.
\end{proof}

\begin{corollary}[scalar prior-quality constant]
\label{cor:ei-scalar}
For scalar trust $\alpha$ and run-fixed prior $\pi$ with miss
$\beta_\star=1-\max_{x^\star\in X^\star}\pi(x^\star)$ at the optimizer,
\begin{equation*}
L_n(\mathrm{EI}^M)\ \le\ \frac{1}{1-\alpha\beta_\star}\,B_n(\mathrm{EI}).
\end{equation*}
\end{corollary}

\begin{proof}
$M(\mathrm{opt})=1-\alpha(1-\pi(x^\star))=1-\alpha\beta_\star$ and $\sup_xM\le1$ in
\eqref{eq:lambda}.
\end{proof}

\begin{corollary}[correct at an optimizer]
\label{cor:ei-free}
If $\pi(x^\star)=1$ at some global optimizer, then $\Lambda(M)=1$: the shaped run satisfies the
\emph{same} theoretical upper bound as unshaped EI, at any trust level, regardless of the belief
elsewhere.  This is a statement about the bound; it does not assert that the realized shaped
trajectory dominates a realized vanilla-EI trajectory.
\end{corollary}

\begin{proof}
$M(\mathrm{opt})=1$ and $\sup_xM\le1$ give $\Lambda=1$ in \eqref{eq:lambda}.
\end{proof}

\begin{corollary}[elicitation--compilation error]
\label{cor:ei-elicitation}
Let $g:\mathcal X\to[0,1]$ be the source's latent belief landscape with miss
$\beta_g=1-g(x^\star)$ at a global optimizer, and suppose the elicitation-plus-compilation pipeline
produces $\pi$ with $\epsilon=\|\pi-g\|_\infty$.  Then for scalar trust $\alpha$,
\begin{equation*}
\Lambda(M)\ \le\ \frac{1}{1-\alpha\,\min(1,\;\beta_g+\epsilon)} .
\end{equation*}
The pipeline enters the constant through two scalars added together---the source's miss at the
optimizer and the sup-norm compilation error---and the penalty never exceeds the ceiling
$(1-\alpha_{\max})^{-1}$.
\end{corollary}

\begin{proof}
$\pi(x^\star)\ge g(x^\star)-\epsilon$ gives $1-\pi(x^\star)\le\beta_g+\epsilon$, and
$1-\pi(x^\star)\le1$ always; insert into \eqref{eq:lambda} via
$M(\mathrm{opt})=1-\alpha(1-\pi(x^\star))$.
\end{proof}

\begin{remark}[time-varying trust]
If the trust schedule satisfies $\sup_x\alpha_t(x)\to0$, the same argument applied at the
pigeonhole iteration shows that the effective constant approaches one asymptotically for any
prior.  Decay is not required for convergence in this noiseless EI regime, the rate is preserved
at any fixed $\alpha\le\alpha_{\max}$.
\end{remark}

\subsection{Regime II: noisy finite-domain raw GP-UCB}
\label{app:theory-ucb}

Back to the maximization convention.

\begin{assumption}[UCB regime]
\label{assump:ucb}
(i)~Finite candidate set $D$ with $|D|=N$; kernel $k$ with $k(x,x)\le1$, fixed during the run,
noise model fixed; (ii)~observations $y_t=f(x_t)+\varepsilon_t$ with noise variance $\sigma_n^2>0$
in the GP update; (iii)~the acquisition is raw GP-UCB,
$U_t(x)=\mu_{t-1}(x)+\sqrt{\beta_t}\,\sigma_{t-1}(x)$ with $\beta_t$ nondecreasing---no
incumbent-relative and no min--max-normalized variant; (iv)~shaped selection
$x_t\in\arg\max_{x\in D}U_t(x)M_t(x)$ with history-measurable $M_t\in[m,1]$;
(v)~\emph{re-querying of candidates is allowed} (no exclusion mask), so the optimum remains
selectable at every step; (vi)~$f(x^\star)\ge0$, a normalization treated as part of the algorithm
(satisfied when responses are nonnegative, or after subtracting a known lower bound; the
multiplicative rule is not shift-invariant, so the shift is part of the method's definition).
Two standard facts are imported.  \emph{Confidence event}: there are standard choices of $\beta_t$
such that with probability at least $1-\delta$,
\begin{equation}
|f(x)-\mu_{t-1}(x)|\ \le\ \sqrt{\beta_t}\,\sigma_{t-1}(x)
\qquad\text{for all }x\in D\text{ and all }t\ge1
\label{eq:conf}
\end{equation}
(for $f$ sampled from the GP, $\beta_t=2\log(N\pi^2t^2/(6\delta))$,
\citealp{DBLP:conf/icml/SrinivasKKS10}; for $f$ in the kernel's RKHS with sub-Gaussian noise, the
self-normalized construction of \citealp{pmlr-v70-chowdhury17a}, valid for any adapted selection
sequence).  Call the event in \eqref{eq:conf} $E$.  \emph{Information-gain bound}: for any
selection sequence, with $\beta_t$ nondecreasing and $k\le1$,
\begin{equation}
\sum_{t=1}^{T}4\beta_t\sigma_{t-1}^2(x_t)\ \le\ C_1\,\beta_T\,\gamma_T,
\qquad C_1=\frac{8}{\log(1+\sigma_n^{-2})},
\label{eq:infogain}
\end{equation}
where $\gamma_T$ is the maximum information gain after $T$ rounds
\citep[Lemma~5.4]{srinivas2009gpucbextended}.
\end{assumption}

\begin{lemma}[finite-domain information gain]
\label{lem:gamma}
On finite $D$ with $k\le1$:
$\gamma_T\le\frac{N}{2}\log\bigl(1+T\sigma_n^{-2}/N\bigr)=O(N\log T)$.
\end{lemma}

\begin{proof}
The information gain of any $T$ observations is
$\frac12\log\det(I+\sigma_n^{-2}K_T)$ with $K_T$ the $T\times T$ kernel matrix of the queried
points.  $K_T$ is positive semidefinite of rank at most $N$ (all columns lie in the span of the
$N$ candidate feature vectors), so it has at most $N$ nonzero eigenvalues $\lambda_i\ge0$ with
$\sum_i\lambda_i=\operatorname{tr}(K_T)\le T$.  Then
$\log\det=\sum_{i\le N}\log(1+\sigma_n^{-2}\lambda_i)
\le N\log\bigl(1+\sigma_n^{-2}(\sum_i\lambda_i)/N\bigr)$ by concavity of $\log(1+u)$ (Jensen),
which is at most $N\log(1+\sigma_n^{-2}T/N)$.
\end{proof}

Define the per-step \emph{misplaced-trust penalty at the optimum}
\begin{equation*}
\theta_t\ :=\ \min_{x^\star\in X^\star}\frac{1-M_t(x^\star)}{M_t(x^\star)}
\ =\ \frac{1}{M_t(\mathrm{opt})}-1,
\qquad 0\le\theta_t\le\frac{1-m}{m},
\end{equation*}
and $\bar\theta_T=\max_{t\le T}\theta_t$.

\begin{theorem}[noisy shaped GP-UCB regret]
\label{thm:ucb}
Under Assumption~\ref{assump:ucb}, on the event $E$, the cumulative regret
$R_T=\sum_{t\le T}\bigl(f(x^\star)-f(x_t)\bigr)$ of shaped GP-UCB satisfies
\begin{equation}
R_T\ \le\ f(x^\star)\sum_{t=1}^{T}\theta_t\ +\ (1+\bar\theta_T)\,\sqrt{C_1T\beta_T\gamma_T}.
\label{eq:ucb-regret}
\end{equation}
The second term is the usual information-gain term of unshaped GP-UCB, inflated by at most
$1+\bar\theta_T\le1/m$; the first, additive term is the cumulative price of misplaced trust and
depends on the belief \emph{only through its value at the optimum}: confident but wrong beliefs
about regions that do not contain the optimum contribute nothing to the bound.
\end{theorem}

\begin{proof}
Fix $t$ and work on $E$.  Let $x^\star$ attain the minimum in $\theta_t$.

\emph{Step 1 (optimism through the multiplier).}  By \eqref{eq:conf}, $U_t(x^\star)\ge
f(x^\star)\ge0$.  By shaped optimality and $M_t>0$,
\begin{equation*}
U_t(x_t)M_t(x_t)\ \ge\ U_t(x^\star)M_t(x^\star)\ \ge\ f(x^\star)M_t(x^\star)\ \ge\ 0,
\end{equation*}
so $U_t(x_t)\ge0$, and
$f(x^\star)\le U_t(x_t)\,M_t(x_t)/M_t(x^\star)\le U_t(x_t)(1+\theta_t)$, using $M_t(x_t)\le1$ and
$1/M_t(x^\star)=1+\theta_t$.

\emph{Step 2 (value at the query).}  By \eqref{eq:conf},
$f(x_t)\ge\mu_{t-1}(x_t)-\sqrt{\beta_t}\sigma_{t-1}(x_t)=U_t(x_t)-2\sqrt{\beta_t}\sigma_{t-1}(x_t)$.

\emph{Step 3 (per-step regret).}  Subtracting,
$r_t\le\theta_t\,U_t(x_t)+2\sqrt{\beta_t}\sigma_{t-1}(x_t)$.  By \eqref{eq:conf} again,
$U_t(x_t)\le f(x_t)+2\sqrt{\beta_t}\sigma_{t-1}(x_t)\le
f(x^\star)+2\sqrt{\beta_t}\sigma_{t-1}(x_t)$, so
\begin{equation*}
r_t\ \le\ \theta_t f(x^\star)+(1+\theta_t)\,2\sqrt{\beta_t}\,\sigma_{t-1}(x_t).
\end{equation*}

\emph{Step 4 (sum).}  Sum over $t\le T$, bound $1+\theta_t\le1+\bar\theta_T$, and apply
Cauchy--Schwarz with \eqref{eq:infogain}:
$\sum_t2\sqrt{\beta_t}\sigma_{t-1}(x_t)\le\sqrt{T\sum_t4\beta_t\sigma_{t-1}^2(x_t)}
\le\sqrt{C_1T\beta_T\gamma_T}$.
\end{proof}

\begin{corollary}[correct at an optimizer]
\label{cor:ucb-free}
If $\pi_t(x^\star)=1$ for all $t$ at some optimizer, then $M_t(\mathrm{opt})=1$, $\theta_t=0$, and
shaped GP-UCB satisfies the \emph{vanilla} GP-UCB bound
$R_T\le\sqrt{C_1T\beta_T\gamma_T}$ on $E$---at any trust level, however wrong the belief is
elsewhere.
\end{corollary}

\begin{proof}
$\theta_t=0$ collapses \eqref{eq:ucb-regret}.
\end{proof}

\begin{corollary}[decaying trust]
\label{cor:ucb-decay}
If $\alpha_t\le\min(\alpha_{\max},\,c/t)$, then $\theta_t\le(c/m)/t$ and, on $E$,
\begin{equation*}
R_T\ \le\ \frac{c\,f(x^\star)}{m}\,(1+\log T)\ +\ \frac1m\,\sqrt{C_1T\beta_T\gamma_T}:
\end{equation*}
the additional cumulative term is $O(\log T)$ for any prior, however adversarial; cumulative
regret is sublinear under the usual information-gain conditions (on finite $D$,
$\beta_T\gamma_T=O(N\log^2T)$ by Lemma~\ref{lem:gamma}); and the best-sampled latent simple regret
vanishes, since $S_T\le R_T/T\to0$.
\end{corollary}

\begin{proof}
$\theta_t=(1-M_t(\mathrm{opt}))/M_t(\mathrm{opt})\le\alpha_t(1-\pi_t(x^\star))/m\le(c/t)/m$;
$\sum_{t\le T}1/t\le1+\log T$; and $1+\bar\theta_T\le1+(1-m)/m=1/m$.  The simple-regret statement
is $\min_{t\le T}r_t\le R_T/T$.
\end{proof}

\begin{proposition}[static misplaced trust can cost linear cumulative regret]
\label{prop:linear}
For every $m\in(0,1)$, there is a two-point problem in the noisy regime of
Assumption~\ref{assump:ucb} and a valid run-fixed multiplier $M\in[m,1]$
such that, on the confidence event $E$, shaped GP-UCB has
\[
R_T=(1-c)T-O(\log T),
\qquad m<c<1.
\]
Thus static shaping can incur linear cumulative regret even with positive
observation noise.
\end{proposition}

\begin{proof}
Let $D=\{a,b\}$, $f(a)=1$, and $f(b)=c$, where $m<c<1$.
Use the fixed kernel $k(x,x')=\mathbf{1}\{x=x'\}$, a zero GP prior mean,
and independent Gaussian observation noise of variance $\sigma_n^2>0$,
with the same variance in the posterior update.  Set $M(a)=m$ and $M(b)=1$.
These multipliers are realized by $\alpha=1-m$, $\pi(a)=0$, and $\pi(b)=1$.
Use the shaped selection rule from the first round and allow re-querying.

The objective belongs to the kernel's RKHS, with
$\|f\|_{\mathcal H_k(D)}=\sqrt{1+c^2}$.
Choose the nondecreasing RKHS confidence schedule from
Assumption~\ref{assump:ucb}; on this fixed two-point domain it can be chosen
with $\beta_t=O(\log(t+1))$ and $\beta_t\to\infty$, since
$\gamma_t=O(\log(t+1))$.
Work on its confidence event $E$.

Write $N_a(t-1)$ for the number of observations of $a$ before round $t$.
The independent-arm kernel gives
\[
\sigma_{t-1}^2(a)
=\frac{\sigma_n^2}{\sigma_n^2+N_a(t-1)}.
\]
Whenever the shaped rule selects $a$, optimality and confidence imply
\[
mU_t(a)\ge U_t(b)\ge c,
\qquad
U_t(a)\le 1+2\sqrt{\beta_t}\,\sigma_{t-1}(a).
\]
Consequently,
\[
N_a(t-1)
\le \frac{4m^2\sigma_n^2\beta_t}{(c-m)^2}-\sigma_n^2.
\]
Using monotonicity of $\beta_t$ and accounting for the last selection of
$a$ (with $N_a(T)=0$ if it is never selected), we obtain
\[
N_a(T)\le 1+\frac{4m^2\sigma_n^2\beta_T}{(c-m)^2}
=O(\log T).
\]
Each query of $b$ incurs regret $1-c$, whereas a query of $a$ incurs zero
regret.  Therefore
\[
R_T=(1-c)\bigl(T-N_a(T)\bigr)=(1-c)T-O(\log T).
\]
\end{proof}

Three comments on Proposition~\ref{prop:linear}.  First, it establishes that
linear cumulative regret can occur within the positive-noise regime of
Theorem~\ref{thm:ucb}.  The linear dependence on $T$ is therefore a real
worst-case possibility, although the construction does not establish
sharpness of the coefficient in the upper bound.  Second,
Corollaries~\ref{cor:ucb-free} and~\ref{cor:ucb-decay} give sufficient
conditions for recovering sublinear cumulative regret: correctness at an
optimizer or decaying trust.  They are not necessary conditions; for example,
a positive multiplier that is constant across candidates leaves the vanilla
GP-UCB selection rule unchanged.  The default \hapbo{} configuration keeps
trust static within a run and accepts the resulting worst-case trade-off,
because deployment campaigns of 8--30 queries are judged by the best
configuration found rather than by cumulative regret.  Third, the proposition
does not imply a failure of discovery.  In this construction, $a$ must
eventually be queried on $E$: otherwise $U_t(a)=\sqrt{\beta_t}\to\infty$,
while repeated observations of $b$ give
$c\le U_t(b)\le c+2\sqrt{\beta_t}\sigma_n/\sqrt{\sigma_n^2+t-1}\to c$,
contradicting the shaped selection rule.  Hence the best-sampled latent
simple regret is zero after the first query of $a$, despite linear
cumulative regret.  This does not establish a recommendation guarantee
from noisy observations.

\end{document}